\documentclass[journal]{IEEEtran}

\usepackage{graphicx}
\usepackage{amsmath}
\usepackage{amsthm}
\usepackage{booktabs}
\usepackage[noend]{algpseudocode}
\usepackage{algorithmicx,algorithm}
\usepackage{color}

\begin{document}
%
\title{Face Deblurring Based on Separable Normalization and Adaptive Denormalization}
%
%
%

\author{Xian Zhang$^{*}$,
        Hao Zhang$^{*}$,
        Jiancheng Lv, ~\IEEEmembership{Member,~IEEE},
       Xiaojie Li$^{\dagger}$,~\IEEEmembership{Member,~IEEE},
\thanks{X. Zhang, H. Zhang, X. Li are with the Chengdu University of Information Technology, China.J. Lv is with the Sichuan University, China.}
\thanks{$^{*}$ Co-first authors}
\thanks{$^{\dagger}$ Corresponding authors: lixj@cuit.edu.cn.}

}

%
%

\markboth{Journal of \LaTeX\ Class Files,~Vol.~14, No.~8, August~2015}%
{Shell \MakeLowercase{\textit{et al.}}: Bare Demo of IEEEtran.cls for IEEE Journals}
%



\maketitle

\begin{abstract}


Face deblurring aims to restore a clear face image from a blurred input image with more explicit structure and facial details. However, most conventional image and face deblurring methods focus on the whole generated image resolution without consideration of special face part texture and generally produce unsufficient details.
Considering that faces and backgrounds have different distribution information, in this study, we designed an effective face deblurring network based on separable normalization and adaptive denormalization (SNADNet).
First, We fine-tuned the face parsing network to obtain an accurate face structure. Then, we divided the face parsing feature into face foreground and background. Moreover, we constructed a new feature adaptive denormalization to regularize fafcial structures as condition of the auxiliary to generate more harmonious and undistorted face structure. In addition, we proposed a texture extractor and multi-patch discriminator to enhance the generated facial texture information.
Experimental results on both CelebA and CelebA-HQ datasets demonstrate that the proposed face deblurring network restores face structure with more facial details and performs favorably against state-of-the-art methods in terms of structured similarity indexing method (SSIM), peak signal-to-noise ratio (PSNR), Frechet inception distance (FID) and L1, and qualitative comparisons.
\end{abstract}


%
\IEEEpeerreviewmaketitle

\begin{figure*}[!h]
    \setlength{\abovecaptionskip}{-0.02cm}
   \setlength{\belowcaptionskip}{-0.3cm}
    \centering
    \includegraphics[width=1.0\textwidth]{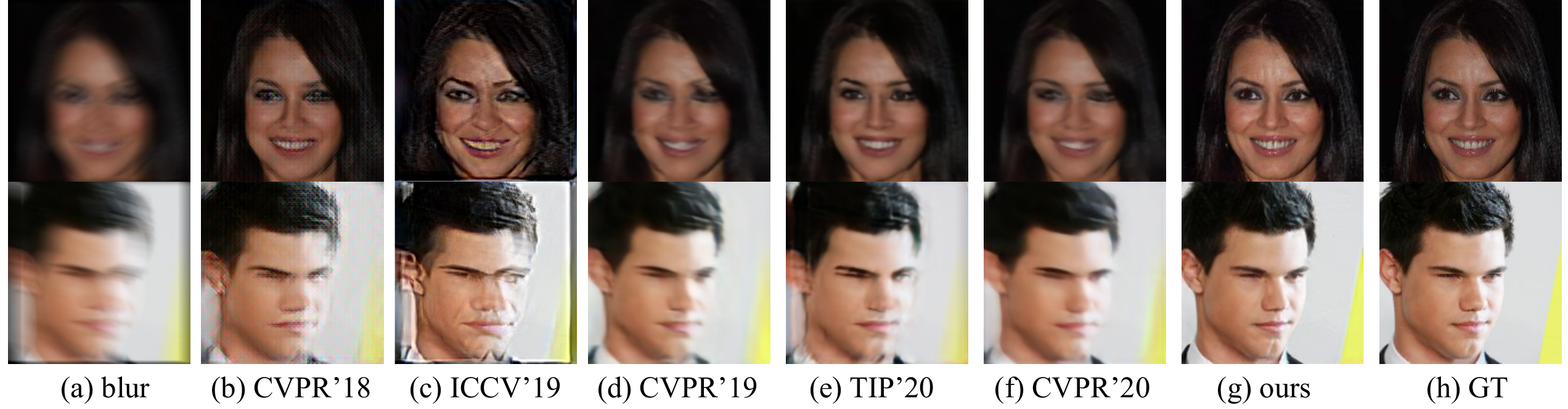}
   \caption{{(a) Blur face, (b) DeblurGAN  (CVPR'18)~\cite{kupyn2018deblurgan}, (c) DeblurGAN-V2 (ICCV'19)~\cite{kupyn2019deblurgan}, (d) DMPHN (CVPR'19)~\cite{zhang2019deep}, (e) UMSN (TIP'20)~\cite{yasarla2020deblurring}, (f) MPRNet (CVPR'21) ~\cite{zamir2021multi}, (g) SNADNet, (h) GT on CelebA dataset. DeblurGAN (CVPR'18) and DeblurGAN-V2 (ICCV'19) produce more unharmonious and distorted face structures. DMPHN (CVPR'19) is over-smooth and not sufficiently sharp. UMSN (TIP'20) cannot produce a sufficiently detailed face texture. MPRNet (CVPR'21) produces large artifacts in the generated faces.}}
    \label{fig_ways}
\end{figure*}

Face deblurring is useful in various fields and applications, including face recognition~\cite{keinert2019robust,chen2021multi}, face alignment~\cite{wu2018look,chang2017faceposenet,wang2019adaptive}, and rapid suspect identification to assist criminal investigations~\cite{yasarla2020deblurring}.
Conventional deblurring methods model the blurring process as the convolution of a clean image with a blur kernel and solve this problem based on the maximum a posterior (MAP) framework. Currently, while the problem is ill-posed, various state-of-the-art algorithms applied various constraints (natural image priors) to model deblurring (i.e., $L_0$ intensity \cite{pan2016l_0} and  dark channel \cite{pan2016blind}) to constrain the solution space. Therefore, face deblurring, which is a special case of image deblurring that aims to restore a clear image from a blurred input image, requires more attention because of its highly structured and special facial components (e.g., eyes and brows).

Although image deblurring~\cite{zhang2021exposure,asim2020blind} has been researched for a long time, it still generates unsatisfactory results with easily detectable flaws in real scenes. Fig.~\ref{fig_ways} illustrated the best deblurring results of several state-of-the-art methods (DeblurGAN (CVPR'18)~\cite{kupyn2018deblurgan}, DeblurGAN-v2 (ICCV'19)~\cite{kupyn2019deblurgan}, DMPHN (CVPR'19)~\cite{zhang2019deep}, UMSN (TIP'20)~\cite{yasarla2020deblurring}, and MPRNet (CVPR'21)~\cite{zamir2021multi}) on a low-quality CelebA dataset with $25,000$ blur kernels.
It is apparent that all SOTA methods are good at deblurring. However, finding ways to generate more accurate facial features and details are ignored, which is considered as a drawback of these techniques.
Fig.~\ref{fig_ways} (a) is the original blurred image. DeblurGAN~\cite{kupyn2018deblurgan}, and DeblurGAN-V2, proposed by kupyn, play an important role in a deblurring image, but they produce more unharmonious and distorted face structure (see Fig.~\ref{fig_ways} (b) and (c)). Zhang et al. proposed a new DMPHN method~\cite{zhang2019deep} but the whole deblurred image is rather smooth and not sharp enough (see Fig.~\ref{fig_ways} (d)).
UMSN, proposed by Yasarla et al~\cite{yasarla2020deblurring} cannot produce a sufficiently detailed face texture (see Fig.~\ref{fig_ways} (e)).
MPRNet, proposed by zamir et al.~\cite{zamir2021multi} in 2021, also produces large artifacts in the generated faces, especially in the eye area (see Fig.~\ref{fig_ways} (f)). This illustrates that these SOTA methods fail to produce more facial details when dealing with narrow edges or highly textured regions in the face. Therefore, new single image deblurring models are necessary to overcome the drawbacks of existing models.

In this paper, we propose an effective face deblurring network based on separable normalization and adaptive denormalization (SNADNet).
First, we fine-tune the face parsing network to achieve a more accurate contour structure of the blurred face. We design a new separable normalization and adaptive denormalization (SNAD) to regularize facial structures and generate more harmonious and undistorted face structures compared to previous studies that consider the contour structure as a prior or specific constraint. In addition, the texture extractor, multi-region reconstruction loss, and multi-patch discriminator are designed to enhance the quality of the generated face image. The main contributions of this study are summarized as follows:
\begin{itemize}
\item We propose SNAD that incorporates face parsing with fine-tuning, produces more accurate face structures by normalizing the foreground and background separably, and makes adaptive denormalization. The purpose of this is to let the network learn face attribute information.

\item We propose a texture extractor to extract face texture information, enhancing the facial texture information in shallow feature maps and producing sharp images.

\item We divide the face image into four regions (facial, back, hair and skin) and design multi-region reconstruction loss with different weights, which concentrate more on facial region and yield more detailed facial texture.

\item We design a multi-patch discriminator learning method that captures image features from multiple receptive fields, which generates a clear face image. In addition, it can be easily applied to other deep models.


\end{itemize}

The paper is organized as follows. Section~\uppercase\expandafter{\romannumeral2} introduces the related image deblurring techniques, face deblurring methods, and common normalization problems in face deblurring. Section~\uppercase\expandafter{\romannumeral3} presents our proposed SNADNet method, In addition,
Section~\uppercase\expandafter{\romannumeral4} discusses the experimental evaluation of our proposed method and compares it with existing state-of-the-art methods. Finally, this study is concluded in Section~\uppercase\expandafter{\romannumeral5}.

\section{Related Work}
\subsection{Blur Kernels}
The blur kernel is a necessary step in obtaining a high-quality image.
Moreover, single-image deblurring aims to recover a clear image $x$ from a single blurred input image $y$ \cite{shen2018deep}. This is modeled as the convolution of a clean image $x$ with a blur kernel $k$.
Formally,
\begin{equation}
\label{equblur}
\begin{aligned}
   y=x\otimes k+\eta
\end{aligned}
\end{equation}
where $\otimes$ is the convolution operation and $\eta$ is the noise. Solving the deconvolution of Eq.(\ref{equblur}) is an ill-posed problem (highly under-constrained) and falls roughly into non-blind deblurring and blind deblurring models. Both $y$ and $k$ are given in the former group, while the latter group aims to estimate $x$ using $y$, which is more consistent with the actual situation and is preferable. Generally, more complicated kernels are difficult to solve. Moreover, in this study, we focused on the latter group as the main research focus. In this section, we introduce image deblurring and face deblurring.

\subsection{Image Deblurring.}
Deep learning has been successfully applied to image deblurring \cite{xu2017learning,tao2018scale,kupyn2018deblurgan,zhang2019deep,kupyn2019deblurgan,li2020efficient,cai2020dark,li2020dynamic}. For example, Xu et al. \cite{xu2017learning} reported that the basic GAN~\cite{goodfellow2014generative,arjovsky2017wasserstein,mao2017least} is effective in capturing semantic information, but the recovered images usually contain content and structural errors. Therefore, they introduced pixel-wise loss and feature matching loss to restore a clear high-resolution image from a blurry low-resolution input. In contrast, Tao et al.~\cite{tao2018scale} showed that directly applying the existing encoder-decoder structure cannot produce optimal results and is not the best choice for deblurring tasks. They downsampled a sequence of blurry images from the input image at different scales and gradually restored the sharp image at different resolutions in a pyramid. Zhang et al. \cite{zhang2019deep} thought that simply increasing the model depth with finer-scale levels and many parameters cannot improve the quality of deblurring, and address the above challenges via a fine-to-coarse hierarchical representation. Moreover, both DeblurGAN~\cite{kupyn2018deblurgan} and DeblurGAN-v2~\cite{kupyn2019deblurgan}, proposed by Kupyn et al., play an important role in image deblurring. They think that L1 or L2, as the pixel-level reconstruction loss, would cause the generated image to be too smooth. Therefore, they proposed the use of perceptual loss~\cite{johnson2016perceptual} for image deblurring. However, directly applying image deblurring methods to face deblurring cannot produce optimal results due to its special particularities (see Fig.~\ref{fig_ways} (b) and (c)].
Zhang et al. ~\cite{zhang2019deep} considered that multi-scale and scale-recurrent network structures were very inefficient in the task of deburring and proposed a deep stacked hierarchical multi-patch network (DMPHN) to deal with multi-scale~\cite{nah2017deep} and scale-recurrent~\cite{tao2018scale} problems. They split the image into several patches and fused them to achieve good results. However, DMPHN does not perform well in face deblurring and generates too smooth faces with many missing details (see Fig.~\ref{fig_ways} (d)). In 2021, Zamir et al. ~\cite{zamir2021multi} proposed a new MPRNet network, which employs multiple stages to complete image deblurring with multiple patches. However, MPRNet has two major drawbacks. First, MPRNet has many parameters, which requires more powerful hardware during training. Second, multiple stages take a long time to complete the training. MPRNet is not a better choice for face deblurring since it generates insufficient details, as shown in Fig.~\ref{fig_ways} (f).

\subsection{ Face Deblurring}
Most existing deblurring algorithms focus on whole image deblurring and not on the quality of the reconstructed facial structure and texture information. Furthermore, as shown in \cite{shen2018deep} and \cite{song2019joint}, face images typically have fewer textures and edges after estimating blur kernels, making it more difficult to restore the structure and texture of the blurred face. Therefore, forcing various priors on the blur kernel or the latent image has been exploited to improve face deblurring results (e.g., dark channel prior \cite{pan2016blind}, semantic priors \cite{shen2018deep}, face parsing \cite{yasarla2020deblurring}). In particular, Shen et al. \cite{shen2018deep} demonstrated face images with high structure and shared several key semantic components (e.g., eyes and mouths). This semantic information can provide a strong prior to restoration. They incorporate global semantic priors as inputs and impose local structure losses to regularize the output within a multi-scale deep CNN.
However, it fails to reconstruct the eyes and mouth regions without reweighting the importance of less represented semantic regions, as shown in \cite{yasarla2020deblurring} (see Fig.~\ref{fig_ways} (e)).
Song et al. \cite{song2019joint} utilizes the domain-specific knowledge of human faces to recover high-quality faces and employs facial component matching to restore facial details.
To address the imbalance of different semantic classes, \cite{yasarla2020deblurring} proposed an uncertainty-guided
multistream semantic networks (UMSNs). It learns class-specific features independently by subnetworks with the reconstruction of a single semantic class and combines them to deblur the entire face image. During training, they incorporated the confidence measure loss to guide the network toward challenging regions of the human face, such as the eyes. However, it failed to restore sharp images with more facial details (see Fig.~\ref{fig_ways} (e)). In the following section, an approach is proposed to overcome this problem.

\begin{figure}[!h]
    \setlength{\abovecaptionskip}{-0.02cm}
   \setlength{\belowcaptionskip}{-0.3cm}
    \centering
    \begin{minipage}[t]{0.48\textwidth}
    \includegraphics[width=1.0\textwidth]{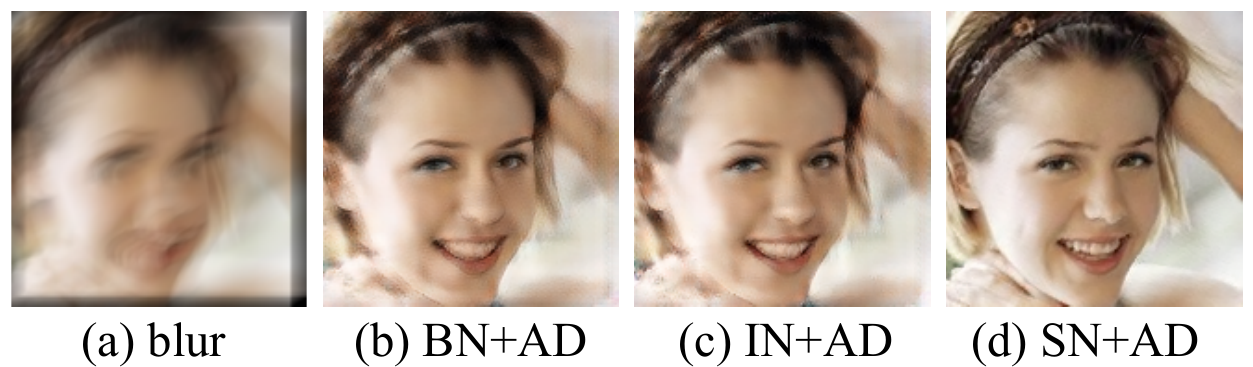}
    \end{minipage}
   \caption{Simulation results were validated by different classical normalizations combined with our proposed adaptive denormalization (AD) to compare and contrast the differences. (a) Blur face, (b) BNAD result (with batch normalization and AD), (c) INAD result (with instance normalization and AD), and (d) SNAD result (with our separable normalization (SN) and AD).}
    \label{different_normalization}
\end{figure}

\subsection{Normalization Problem}
In deep learning, various normalization techniques (i.e., batch normalization (BN)~\cite{ioffe2015batch}, group normalization (GN)~\cite{wu2018group}, instance normalization (IN)~\cite{ulyanov2016instance}, and layer normalization (LN)~\cite{ba2016layer}) are designed to speed up training and improve network performance. As shown in Fig.~\ref{different_normalization}, we illustrate the deblurred results obtained by BN and IN combined with our proposed AD. Compared with our separable normalization (SN) (see Fig.~\ref{different_normalization} (d)), it is known that BN and IN lead to ambiguous results in face deblurring tasks (see Fig.~\ref{different_normalization} (b) and (c)]. Furthermore, the drawbacks and various problems of normalization are analyzed in the appendix (see Appendix \ref{nor-problem}).

\section{Proposed Method}

\begin{figure*}[t]
    \setlength{\abovecaptionskip}{-0.02cm}
   \setlength{\belowcaptionskip}{-0.3cm}
    \centering
    \includegraphics[width=1.0\textwidth]{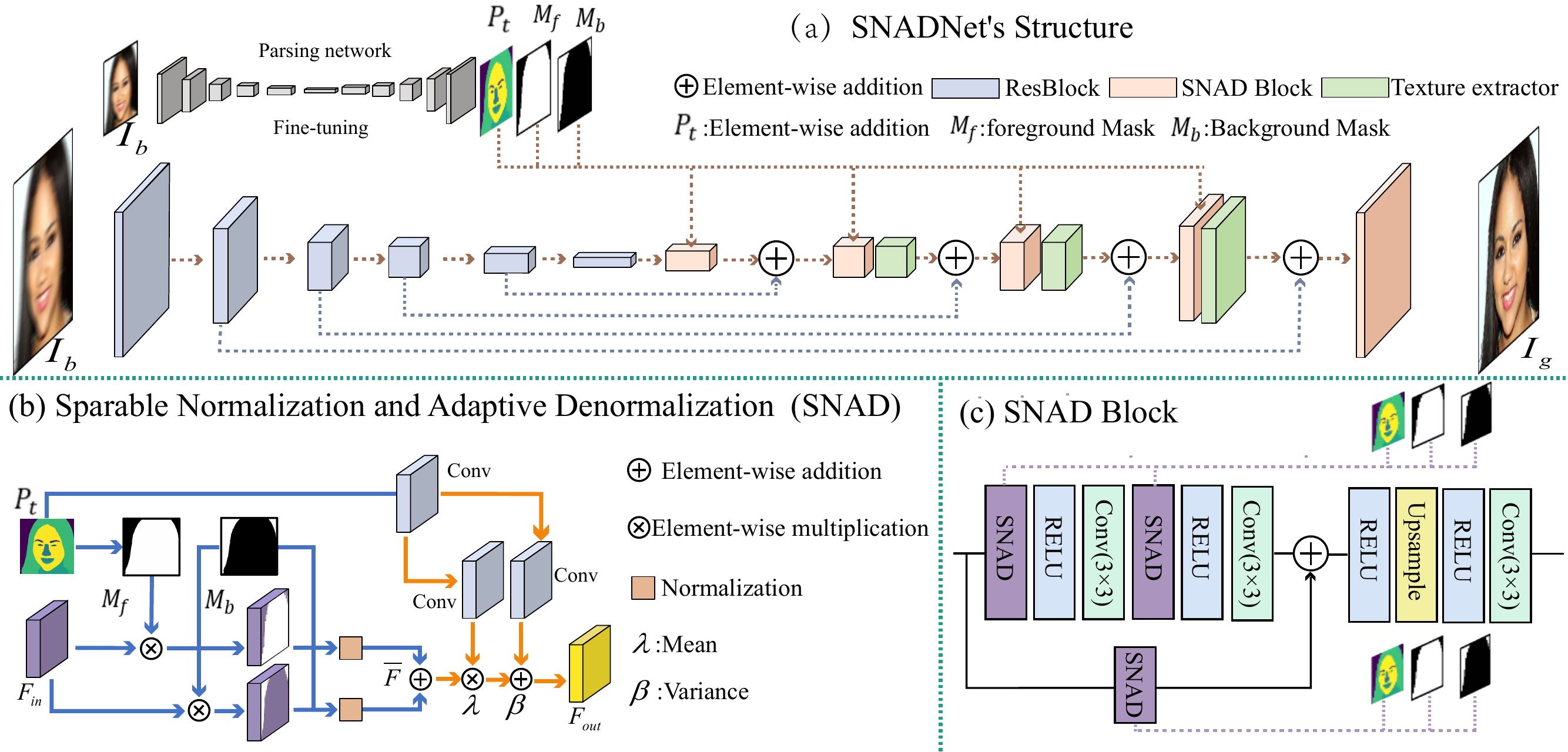}
    \caption{{Overview of the SNADNet architecture, it mainly consists of three modules: face parsing network, Generator, and multi-patch discriminator. Fine-tuning face parsing networks aim to acquire more accurate facial structures from blurred faces. The generator is composed of an encoder and decoder, and mainly includes a Resblock, SNAD block, and texture extractor. We designed an SNAD block to normalize the face foreground and background regions and construct an AD to accurately reconstruct the face structure. A texture extractor is used to extract more accurate facial texture information to enhance the generated texture information.}}
    \label{fig_network}
\end{figure*}

An overview of SNADNet is presented in Fig.~\ref{fig_network}. It consists of two stages. The first stage includes a parsing network, and we used focal loss~\cite{lin2017focal} to fine-tune the face parsing network to extract more accurate face structure information from the blurred face. The second stage includes the generator and the discriminator. Here, we designed separable normalization to normalize the extracted face structure and propose adaptive denormalization (SNAD) to make adaptive denormalization, which guides the generation of the face structure and yields a more symmetrical and harmonious face structure. Moreover, we designed a textural extractor to enhance the face texture and propose a multi-patch discriminator to supervise the generation of a clear face. In the following section, we introduce its details from three parts: parsing network, generator, and discriminator.

\subsection{Face Parsing}
To capture more accurate face structure information from the blurred face image  $x_{blur}$ (see $P_t$ in Fig.~\ref{fig_network}), we fine-tuned the parsing network to acquire a more accurate facial structure.
Because five sense attributes (e.g., facial parts, skin, hair, and background) of face segmentation are very unbalanced, we employ focal loss~\cite{lin2017focal} to fine-tune the parsing network to improve face semantic segmentation. Formally,
\begin{equation}
\label{focal_loss}
\begin{aligned}
    FL=-\alpha_{t}(1-P_{t})^{\gamma }log(P_{t})
\end{aligned}
\end{equation}
where $\alpha_{t}$ is a weighting factor, $P_{t}$ denotes the parsing network prediction, and $\gamma$ is the tunable focusing
parameter, and set $\gamma=2.0$ in our experiment. The backbone code of the face parsing network is available for download on github\footnote{For the face parsing network, we used the released code from https://github.com/zllrunning/face-parsing.PyTorch.}.

\subsection{Generator}
\begin{figure}[t]
\setlength{\abovecaptionskip}{-0.05cm}
    \begin{center}
        \begin{minipage}[t]{0.4\textwidth}
            \includegraphics[width=\linewidth]{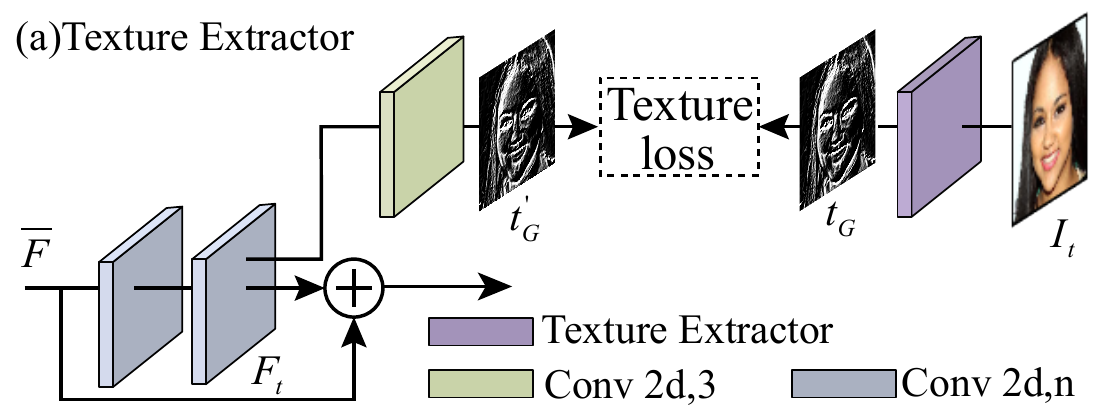}
        \end{minipage}
    \end{center}
    \vspace{-0.1cm}
    \caption{Texture extractor. It extracts the facial texture information from the target face $I_t$. We design a texture extractor as a Laplace operator to extract the information of RGB three channels through depthwise separable convolution and then impose $L_1$ constraints on the learned texture information to enhance the generated texture information.}
    \label{fig_extractor}
    \vskip -0.25in
\end{figure}

Our generator with LinkNet~\cite{chaurasia2017linknet} as the backbone is composed of an encoder and decoder, and mainly includes ResBlock, SNAD block, and texture extractor. We designed the ResBlock with two residuals. As shown in Fig.~\ref{fig_network} (a), a series of ResBlocks make up the encoder and extract both shallow and deep features to obtain key information from blurry faces $I_{b}$. The downsampling was implemented using a convolutional layer with a stride of 2.
The decoder includes an SNAD block and texture extractor (see Fig.~\ref{fig_network} (c) and Fig.~\ref{fig_extractor}).
The SNAD block is used to normalize the face foreground and background regions, and performing AD achieves the effect of accurate reconstruction of the face structure. We designed a texture extractor to extract detailed facial texture information and applied it to the corresponding feature map to reconstruct the face with more detailed and rich texture information.
In the following sections, the SNAD block and texture extractor are described in detail.

s\subsection{SNAD Block}
SNAD (Separable Normalization and Adaptive Denormalization) Block is shown in Fig.~\ref{fig_network} (c). It consists of the SNAD layer, RELU, Conv, and Upsample layers.
Each SNAD layer mainly includes the SN and AD (see Figure~\ref{fig_network} (b)).

{\bf Separable Normalization (SN).}
Inspired by the image inpainting method region normalization (RN) and the previous normalization problem (see Section \ref{nor-problem}), we designed separable normalization to normalize the face foreground and background, which normalized the face foreground and background independently and did not interfere with each other. As shown in Fig.~\ref{fig_network} (b), we divide $P_t$ into face foreground $M_f$ and background $M_b$, which is similar to RN (dividing the image into mask and unmasking two regions) but very different.
Specifically, RN is the batch normalization of the mask and unmask regions in the (B, W, H) dimensions, which is not reasonable for face deblurring. Because the position and shape of each image's face in a batch are not uniformly fixed, the normalization along with channel C in a batch will make the normalized content include both the foreground and background regions between different images.
Furthermore, from Section \ref{nor-problem}, we know that conventional BN, GN, IN, and LN normalization lead to a large amount of normalization bias for the face.
Importantly, in contrast to BN, GN, IN, LN, etc., our SN abandons affine transformation during normalization and uses $P_{t}$ to learn the mean and variance of the pixel level, and simultaneously completes the structural constraint of the face through our AD. The AD is discussed in detail in the next section.
Therefore, we redesigned the approach of normalizing the foreground and background regions.

First, we divide the semantic segmentation label $P_t$ tuned from parsing network into both foreground mask $M_f$ and background mask $M_b$. Suppose $(B, C, H, W)$ are the shape of input feature $F_{in}$, then the mean and variance of the foreground can be calculated as follows:
\begin{equation}
\label{F_f}
\begin{aligned}
   F_f=F_{in}*M_{f}
\end{aligned}
\end{equation}

\begin{equation}
\label{mean_f}
\begin{aligned}
   \mu_{f}^{b,c}=\frac{1}{\sum_{m_{f}^{b,c}\in M_{f}}m_{f}^{b,c}}\sum_{b,c,h=1,w=1}^{H,W}F_f^{b,c,h,w}
\end{aligned}
\end{equation}

\begin{equation}
\label{variance_f}
\begin{aligned}
   \sigma_{f}^{b,c}=\sqrt{\frac{1}{\sum_{m_{f}^{b,c}\in M_{f}}m_{f}^{b,c}}\sum_{b,c,h=1,w=1}^{H,W}(F_f^{b,c,h,w}-\mu_f^{b,c})^{2}+\epsilon }
\end{aligned}
\end{equation}
where $b, c$ represent the $b$th feature and the $c$th channel in batch $B$, respectively. $(h,w)$ represents spatial location. Then, the normalization of the foreground feature is defined as:
\begin{equation}
\label{norm_f}
\begin{aligned}
   \bar{F}_{f}^{b,c}=\frac{1}{\sigma_{f}^{b,c}}(F_{f}^{b,c}-\mu_{f}^{b,c})^{2}
\end{aligned}
\end{equation}
Similarly, the background normalization $\bar{F}_{b}^{b,c}$ is defined in the same manner. Finally, our SN is defined as:
\begin{equation}
\label{concat}
\begin{aligned}
   \bar{F}=\bar{F}_{f}+\bar{F}_{b}
\end{aligned}
\end{equation}
where $\bar{F}$ is the feature obtained after separable normalization. $\bar{F}_{b}^{b,c}$ is the $b$th feature and the $c$th channel in batch $\bar{F}_{b}$. Similarly, the relationship holds between $\bar{F}_{f}^{b,c}$ and $\bar{F}_{f}$.

As shown in Fig.~\ref{fig_network} (b), $\bar{F}$ is obtained after SN. In appendix A, Eq.(\ref{norm}) illustrates that normalizing the face foreground and background regions by SN not only makes both regions follow a standard normal distribution but also makes the whole image follow the standard normal distribution.
In this way, both the foreground and background regions cannot affect each other in the normalization process, and it is also a preparation for the next step of better denormalization.
In particular, our SN normalizes both face foreground and background regions such that the following AD can accurately learn the attribute information of the face.

{\bf Adaptive Denormalization (AD).}
To reconstruct more detailed face content and recover more symmetrical facial structure information, we designed an AD. Specifically, we first fed the semantic label $P_{t}$ obtained by parsing the network with fine-tuning into two parallel convolution layers to learn its pixel-level mean $\gamma$ and standard deviation $\beta$ using Eq.(\ref{equation_m_v}).

\begin{equation}
\label{equation_m_v}
\begin{aligned}
    & \gamma =conv_{2}(conv_{1}(P_{t}))\\
    & \beta =conv_{3}(conv_{1}(P_{t}))\\
\end{aligned}
\end{equation}
Then, the output feature $F_{out}$ is obtained by feature AD:
\begin{equation}
\label{denormalization}
\begin{aligned}
    F_{out} =\bar{F}\otimes \gamma+\beta
\end{aligned}
\end{equation}
where $\otimes$ denotes multiplication by the elements. In this way, facial information can be restored structurally and prevent damage to the loss of facial information.

\begin{figure*}[t]
    \setlength{\abovecaptionskip}{-0.02cm}
   \setlength{\belowcaptionskip}{-0.3cm}
    \centering
    \includegraphics[width=1.0\textwidth]{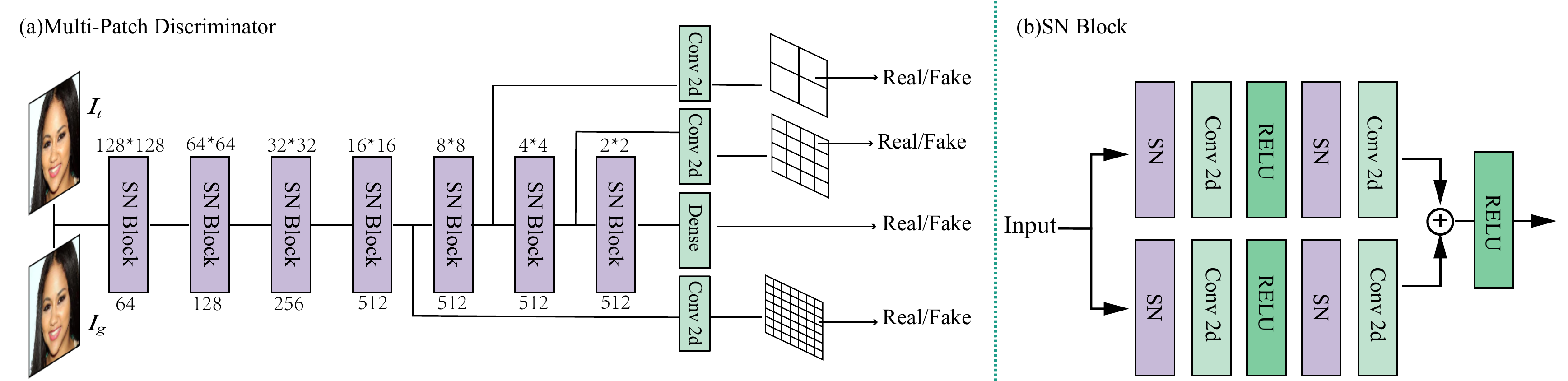}
    \caption{{Overview of the Multi-patch discriminator architecture. (a) Overall framework of multi-patch discriminator. (b) SN block. It consists of two branches in parallel and includes the SN, convolution, and ReLU activation functions. The kernel sizes of both convolutional branches are $4 \times 4$ and $3 \times 3$, respectively. Using two branches in parallel can increase the receptive field.} }
    \label{fig_discriminator_multipatch}
\end{figure*}

Since facial parts, skin, hair, and background have different attributes, they should have their own mean and variance to represent their own information. In contrast to the VAE sampling method, we first normalized the feature to the standard normal distribution with SN, then used face parsing semantic segmentation labels to enable the network to learn the pixel-level mean and variance, and regularize the generated face structure to learn the attribute information of their respective regions by inverse normalization. In this way, the network can learn each attribute information of the face and achieve the purpose of accurate reconstruction of the face structure.

The entire structure of the SNAD block is shown in Fig.~\ref{fig_network} (c)]. Specifically, the previous feature maps ($P_t$ and $F_{in}$) are successively fed into the SNAD, activation function (RELU), and convolution twice to obtain a new middle feature map. Concatenating another output of SNAD, we obtain the final middle feature map, which is then sampled twice the size by the activation function, deconvolution and convolution, and then output. The entire SNAD algorithm process has been introduced in detail in Algorithm ~\ref{algorithmic}.


\begin{algorithm}
\caption{Process of SNAD layer}
\begin{algorithmic}[1]
\Require The set of feature map for current batch $F_{in}$
\Ensure Reconstructed feature map $F_{out}$
\State \textbf{Sparable normalization}
\State Split $P_{t}$ into $M_{f}$ and $M_{b}$
\State Send $F_{in}$ and $M_{f}$ into Eq.(\ref{F_f}) to get the face foreground region $F_{f}$
\State Send $F_{in}$ and $M_{b}$ into Eq.(\ref{F_f}) to get the background region $F_{b}$
\State $M_{f}$ and $F_{f}$ are sent into Eq.(\ref{mean_f}) and Eq.(\ref{variance_f}) to get $\mu^{b,c}_{f}$ and $\sigma^{b,c}_{f}$
\State $M_{b}$ and $F_{b}$ are sent into Eq.(\ref{mean_f}) and Eq.(\ref{variance_f}) to get $\mu^{b,c}_{b}$ and $\sigma^{b,c}_{b}$
\State Use Eq.(\ref{norm_f}) to get foreground normalization region $\bar{F}^{b,c}_{f}$ and background normalization region $\bar{F}^{b,c}_{b}$
\State send $\bar{F}^{b,c}_{f}$ and $\bar{F}^{b,c}_{b}$ Eq.(\ref{norm}) to get $\bar{F}$
\State \textbf{End sparable normalization}
\State \textbf{Adaptive denormalization}
\State Send $P_{t}$ into Eq.(\ref{equation_m_v}) to get adaptive mean $\gamma$ and variance $\beta$
\State Send $\bar{F}$, $\gamma$, and $\beta$ into Eq.(\ref{equation_m_v}) to get denormalization feature $F_{out}$
\State \textbf{End adaptive denormalization}
\State Return $F_{out}$
\end{algorithmic}
\label {algorithmic}
\end{algorithm}

\subsection{Texture Enhancement (TE)}
Liu et al.~\cite{liu2020rethinking} reported in rethinking image inpainting that the shallow feature map of an encoder could extract the texture information of an image, while the deep feature map extracts the structure information of an image. Similarly, we found that the deep feature map of a decoder performs structural restoration when performing information reconstruction, while its shallow feature map performs texture restoration.
Therefore,
we designed a texture extractor to extract more accurate facial texture information from the target image $I_{t}$ and to further enhance the texture information of the shallow feature map. As shown in Fig.~\ref{fig_extractor}, the texture extractor uses the Laplace operator as the convolution kernel, extracts the face texture from the R, G, and B channels of the image $I_{t}$ through depthwise convolution, and obtains the texture map $t_G$. Moreover, the feature map $\bar{F}$ in the generator is sent into two convolution layers and then split into two branches. The first branch predicts the texture of face $t^{'}_{G}$, and the second branch obtains the output from the residual structure. Specifically, $t^{'}_{G}$ and $t_G$ are constrained by $L_{1} $ loss. Then, the final output is obtained through $\bar{F}+F_{t}$ and sent to the next layer. A texture extractor is used to complete the extraction of texture information and is applied in the shallow feature map to ensure that the generator can reconstruct the detailed information of the face and generate a more high-quality face.

Therefore, the generator is composed of the ResBlock, which is used to extract key information from blurry faces $I_{b}$, the SNAD block, which is used to normalize the face foreground and background regions and perform AD to achieve accurate reconstruction of the face structure, and the texture extractor, which is used to extract the face texture information and apply it to the shallow feature map to reconstruct the face with more detailed and rich texture information.


\subsection{Multi-patch Discriminator}
Certain GAN-based methods (e.g., GLCIC~\cite{iizuka2017globally} and LOAHNet~\cite{zhou2020learning}) using multiple discriminators can generate better images but require a large number of parameters. We constructed a multi-patch discriminator with different receptive fields and integrate one global discriminator to supervise the generator and effectively
supervise the generator.
As shown in Fig. ~\ref{fig_discriminator_multipatch} (a), it includes a series of SN (spectral normalization) blocks to stabilize the discriminator.
Moreover, a $3\times3$ convolution is followed by the last three blocks and output features are set with $8\times8$, $4\times4$, and $2\times2$, respectively. The values of each position in the different feature maps represent the perception results obtained under different receptive fields. The last layer of the discriminator is fed into a dense layer to determine whether the entire image is real or fake. The SN block consists of two branches, including SN, convolution, and ReLU activation functions, as shown in Fig.~\ref{fig_discriminator_multipatch} (b). The kernel sizes of the first and second convolutional branches are $4 \times 4$ and $3 \times 3$, respectively. Note that we employed two branches in parallel to increase the receptive field. With this, our multi-patch discriminator could effectively supervise the generator with multiple discriminators, reduce the number of discriminator parameters, and save training cost and time.

\subsection{Loss Function}
In the following section, the multi-region reconstruction loss, texture loss, and adversarial loss are described. The loss of the face parsing network is described by Eq.(\ref{focal_loss}).

\begin{figure}[!h]
    \setlength{\abovecaptionskip}{-0.02cm}
   \setlength{\belowcaptionskip}{-0.3cm}
    \centering
    \begin{minipage}[t]{0.48\textwidth}
    \includegraphics[width=1.0\textwidth]{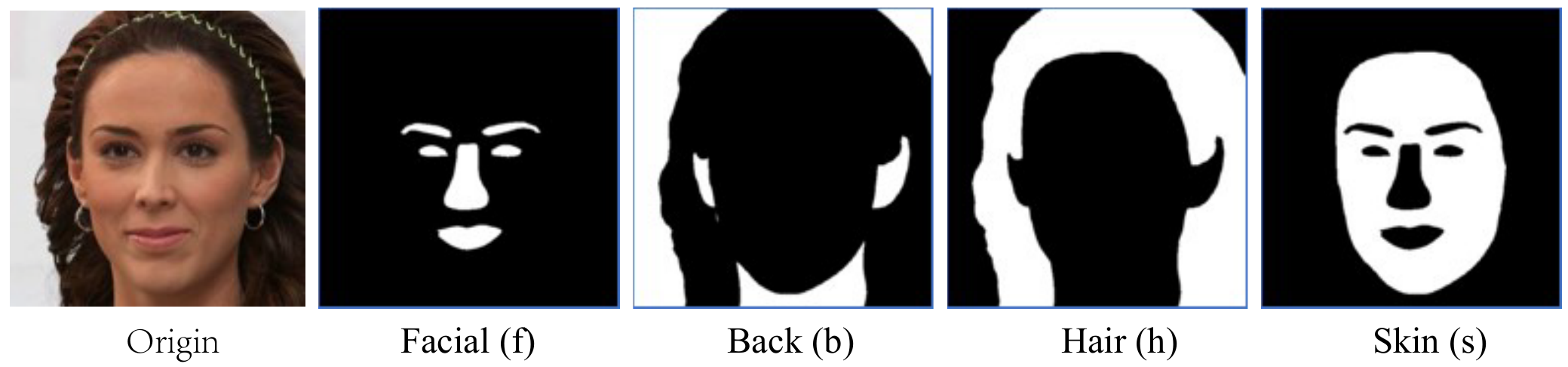}
    \end{minipage}
   \caption{{Four types of mask as ground truth: (a) Original image; (b) Facial mask ($f$);  (c) Back mask ($b$); (d) Hair mask ($h$); (e) Skin mask ($s$).}}
    \label{fig_mask}
\end{figure}

{\bf Multi-region reconstruction loss.}
In contrast with some methods that directly use the $L_1$ loss constraint on the entire image, our SNADNet pays more attention to face reconstruction. Therefore, we designed a multi-region reconstruction loss. We divided the face into facial mask $f$, back mask $b$, hair mask $h$, and skin mask $s$ using a face parsing network with fine-turning, which is inspired by the reconstruction losses of UMSN, as shown in Fig.~\ref{fig_mask}. Subsequently, we constructed four regions of reconstruction loss to restore these four regions.
Formally,

\begin{equation}
\label{recloss}
\begin{aligned}
    loss_{x}^{rec}&=\frac{1}{N}\sum_{i=1}^{N}\frac{1}{sum(x_i)}\left \| x_{i}*I_{g}- x_{i}*I_{t}\right \|
\end{aligned}
\end{equation}
where $I_{g}$ is the generated face by the generator, $I_ {t}$ is the target face, $x\in\{s,f,h,b\}$, and $N$ denotes the number of images in a batch.
The total reconstruction loss is shown in Eq.(\ref{recloss_all}):
\begin{equation}
\label{recloss_all}
\begin{aligned}
    loss^{rec}=\lambda_{1}^{rec}* loss_{s}^{rec}+\lambda_{2}^{rec}*loss_{f}^{rec}\\
    +\lambda_{3}^{rec}*loss_{h}^{rec}+\lambda_{4}^{rec}*loss_{b}^{rec}
\end{aligned}
\end{equation}
where $\lambda_{1}^{rec}=12$, $\lambda_{2}^{rec}=10$, $\lambda_{3}^{rec}=8$, and $\lambda_{4}^{rec}=6$ were set in our experiments. The purpose of setting the super parameter in this manner is to give the highest attention to the face and the lowest attention to the background. The multi-region reconstruction loss could force the generator to focus on the facial structure generation and texture construction.

{\bf Texture loss.} The generator of SNADNet predicts the face texture information in the decoder, as shown in Fig.~\ref{fig_network}. In Fig.~\ref{fig_extractor}, we constrained the predicted texture information $t^{'}_{G}$ with the texture information extracted by the texture extractor, and the texture loss can be expressed as:

\begin{equation}
\label{loss_texture}
\begin{aligned}
    loss^{texture}=\sum_{n=1}^{N=3}\left \| t_{G}-t^{'}_{G} \right \|^{(n)}
\end{aligned}
\end{equation}
where $t_{G}$ is the texture information extracted from the target face $I_{t}$ and $t_{G}^{'}$ is predicted by the network.

{\bf Adversarial loss.}
Our multi-patch discriminator outputs three different patch results and one global result. For the generator's adversarial loss, the Relativistic adversarial loss~\cite{jolicoeur2018relativistic,jolicoeur2018rfdiv}, is used in Eq.(\ref{g_adv}):
\begin{equation}
\label{g_adv}
\begin{aligned}
    loss_{gk}^{adv} &=-E_{(I_g,I_t)\sim (P,Q)}[log(sigmoid(D_{Pk}(I_g)-D_{Pk}(I_t)))]\\
    loss_{g4}^{adv} &= -E_{(I_g,I_t)\sim (P,Q)}[log(sigmoid(D_{G}(I_g)-D_{G}(I_t)))]\\
\end{aligned}
\end{equation}
where $k\in\{1,2,3\}$ and $D_{P1},\cdots, D_{P3}$ denotes patch results with different receptive fields, and $D_{G}$ is the global discriminator. Thus the adversarial loss of generator can be written as:
\begin{equation}
\label{adv_g_loss_all}
\begin{aligned}
    loss_{g}^{adv} = \lambda_{4}^{adv}*loss_{g4}^{adv}+\lambda_{3}^{adv}*loss_{g3}^{adv}\\
    +\lambda_{2}^{adv}*loss_{g2}^{adv}+\lambda_{1}^{adv}*loss_{g1}^{adv}
\end{aligned}
\end{equation}
where set $\lambda_{4}^{adv}=1.0$, $\lambda_{3}^{adv}=0.8$, $\lambda_{2}^{adv}=0.4$, $\lambda_{1}^{adv}=0.2$ in our experiments. We set different weights according to different receptive fields, and the larger the receptive field, the greater the weight.

Finally, the total loss of the generator is shown in Eq.(\ref{loss_all}):
\begin{equation}
\label{loss_all}
\begin{aligned}
    loss_{g}^{all}=\lambda_1*loss^{rec}+\lambda_2*loss_{g}^{adv}+loss^{texture}
\end{aligned}
\end{equation}
where set $\lambda_1=120$ and $\lambda_{2}=0.1$ in our experiments.

Similar to the adversarial loss of the generator, the discriminator's adversarial loss also includes three patch results with different receptive fields and one global result. The difference is that the discriminator's adversarial loss is opposite to the generator's adversarial loss target. Formally,

\begin{equation}
\label{adv_d_loss}
\begin{aligned}
\small
    loss_{dk}^{adv} &= -E_{(I_g,I_t)\sim (P,Q)}[log(sigmoid(D_{Pk}(I_t)-D_{Pk}(I_g)))]\\
    loss_{d4}^{adv} &= -E_{(I_g,I_t)\sim (P,Q)}[log(sigmoid(D_{G}(I_t)-D_{G}(I_g)))]
\end{aligned}
\end{equation}
where $k\in\{1,2,3\}$ and $D_{P1},\cdots, D_{P3}$ denotes patch results with different receptive fields, and $D_G$ is the global discriminator.
The adversarial loss of the entire multi-patch discriminator can be written as:

\begin{equation}
\label{adv_d_loss_all}
\begin{aligned}
    loss_{d}^{adv} = \lambda_{4}^{adv}*loss_{d4}^{adv}+\lambda_{3}^{adv}*loss_{d3}^{adv}\\
    +\lambda_{2}^{adv}*loss_{d2}^{adv}+\lambda_{1}^{adv}*loss_{d1}^{adv}
\end{aligned}
\end{equation}
where $loss_{d}^{adv}$ and $loss_{g}^{adv}$ have the same value in super parameter.
\begin{figure*}[t]
    \setlength{\abovecaptionskip}{-0.02cm}
   \setlength{\belowcaptionskip}{-0.3cm}
    \centering
    \includegraphics[width=1.0\textwidth]{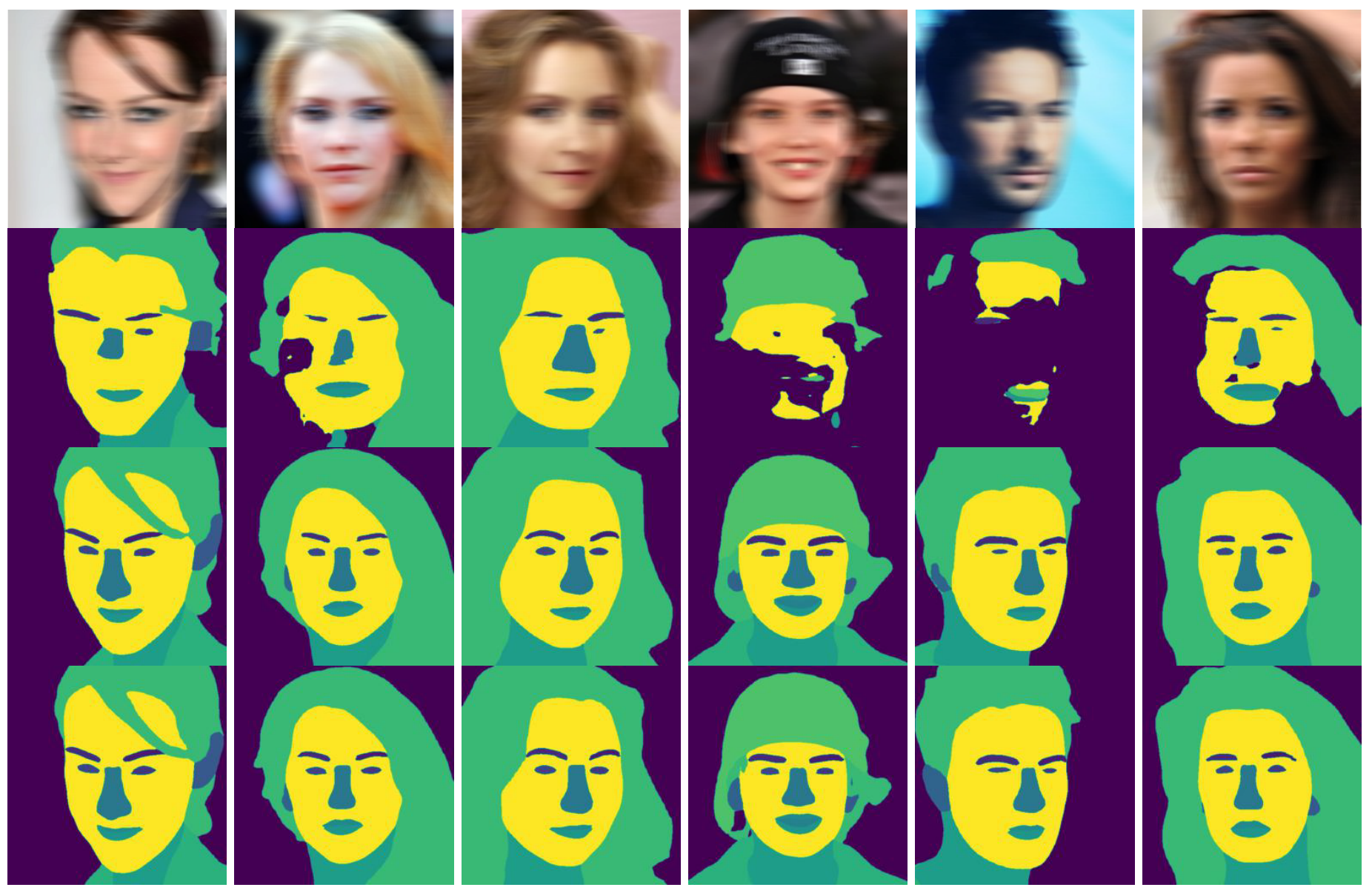}
    \caption{Semantic maps generated by face parsing network from blurry images on the CelebA-HQ dataset. The first row shows the blurry images. The second row consists of the corresponding semantic masks obtained from the face parsing network on the blurry faces without fine-tuning. The third row consists of the corresponding semantic masks obtained from the face parsing network with fine-tuned blurry images. The fourth row consists of the corresponding semantic masks using a face parsing network with fine-tuned clean images.}
    \label{fig_parsing}
\end{figure*}

\section{Experiments}
\subsection{Experimental Settings}
{\bf Datasets:} To investigate the effectiveness of our method, two public face image datasets are used: high-quality  CelebA-HQ (256$\times$256)~\cite{karras2017progressive} and low-quality CelebA (160$\times$160)~\cite{liu2015faceattributes}. Specifically, we divided CelebA-HQ into training set $28,000$, validation set $1,000$ , and test set $1,000$. To simulate a real blur scene, we use a fuzzy kernel with a size of $25$ to blur the motion of the original image with a motion angle of $45$ degrees. CelebA is divided into training set $162,770$, validation set $19,867$, and test set $19,962$. To further demonstrate the performance of our network, we applied a more complex blur to CelebA dataset. Specifically, we generate $25,000$ blur kernels sizes ranging from $13 \times 13$ to $29 \times 29$ , using 3D camera trajectories~\cite{boracchi2012modeling}. Patches of size $160 \times 160$ are extracted from those images and convolved with $2,500$ blur kernels randomly to generated millions of pairs of clean-blurry data. We added Gaussian noise with $\sigma=0.03$ to the blurry images.
Moreover, we construct the ground-truth of the face semantic segmentation labels from both clear datasets by the face parsing network (see Fig.~\ref{fig_mask}).

Furthermore, to illustrate that our generated semantic maps by the face parsing network with fine-tuning and blurry images as input can accurately capture the face structure information, Fig.~\ref{fig_parsing} shows some $P_t$ results without fine-tuning or with different inputs. The first row contains original blurry images. The second row consists of the semantic mask $P_t$ obtained from the face parsing network on the blurry faces without fine-tuning. The third row consists of the corresponding semantic masks $P_t$ by the face parsing network with fine-tuning and blurry images. The fourth row consists of the corresponding semantic mask $P_t$ obtained from the face parsing network with clean images as input. Compared with the second row and fourth row, it is easy to see that our fine-tuning face parsing network on the blurry face can accurately capture the structure information of the face.

{\bf Training Details.} Our network training consists of two stages. In Stage-I, we employ focal loss and SGD optimizer with an initial learning rate of $0.02$, $momentum=0.9$, $weight\_decay=0.0005$ to fine-tune the face parsing network. After the pre-trained face parsing network converges, the deblurring process would be trained in stage-II. In stage-II, Adam optimizer is used with an initial learning rate of $0.0002$, $\beta_{1}=0.5$, $\beta_{2}=0.999$. All experiments are implemented on the Pytorch framework with 12G NVIDIA RTX 2080Ti GPU.

\subsection{Experimental Results}
We compared our method (SNADNet) with five state-of-the-art methods: DeblurGAN (CVPR'18)~\cite{kupyn2018deblurgan}, DeblurGAN-v2 (ICCV'19)~\cite{kupyn2019deblurgan}, DMPHN (CVPR'19)~\cite{zhang2019deep}, UMSN (TIP'20)~\cite{yasarla2020deblurring}, and MPRNet (CVPR'21)~\cite{zamir2021multi}.

\begin{equation}
\label{adv_d_loss_all}
\begin{aligned}
    loss_{d}^{adv} = \lambda_{4}^{adv}*loss_{d4}^{adv}+\lambda_{3}^{adv}*loss_{d3}^{adv}\\
    +\lambda_{2}^{adv}*loss_{d2}^{adv}+\lambda_{1}^{adv}*loss_{d1}^{adv}
\end{aligned}
\end{equation}
where $loss_{d}^{adv}$ and $loss_{g}^{adv}$ have the same value in super parameter.
\begin{figure*}[t]
    \setlength{\abovecaptionskip}{-0.02cm}
   \setlength{\belowcaptionskip}{-0.3cm}
    \centering
    \includegraphics[width=1.0\textwidth]{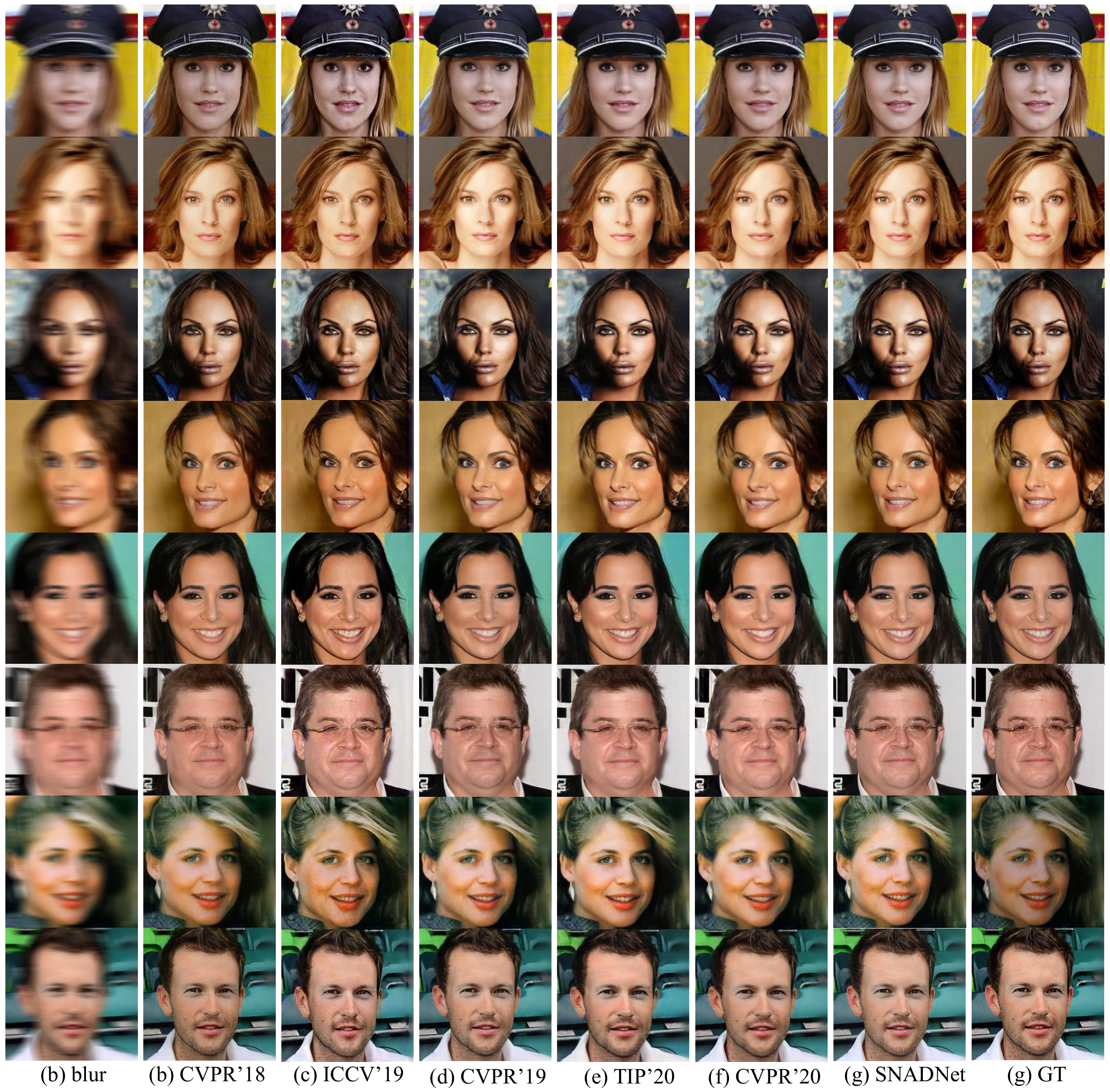}
     \caption{{Qualitative comparisons on the CelebA-HQ dataset. (a) blur face, (b) DeblurGAN,  (c) DeblurGAN-V2, (d) DMPHN, (e) UMSN, (f) MPRNet, (g) SNADNet, and (h) GT. More results are shown in the appendix.}}
     \label{fig_celeba_hq}
\end{figure*}

\begin{equation}
\label{adv_d_loss_all}
\begin{aligned}
    loss_{d}^{adv} = \lambda_{4}^{adv}*loss_{d4}^{adv}+\lambda_{3}^{adv}*loss_{d3}^{adv}\\
    +\lambda_{2}^{adv}*loss_{d2}^{adv}+\lambda_{1}^{adv}*loss_{d1}^{adv}
\end{aligned}
\end{equation}
where $loss_{d}^{adv}$ and $loss_{g}^{adv}$ have the same value in super parameter.
\begin{figure*}[t]
    \setlength{\abovecaptionskip}{-0.02cm}
   \setlength{\belowcaptionskip}{-0.3cm}
    \centering
    \includegraphics[width=1.0\textwidth]{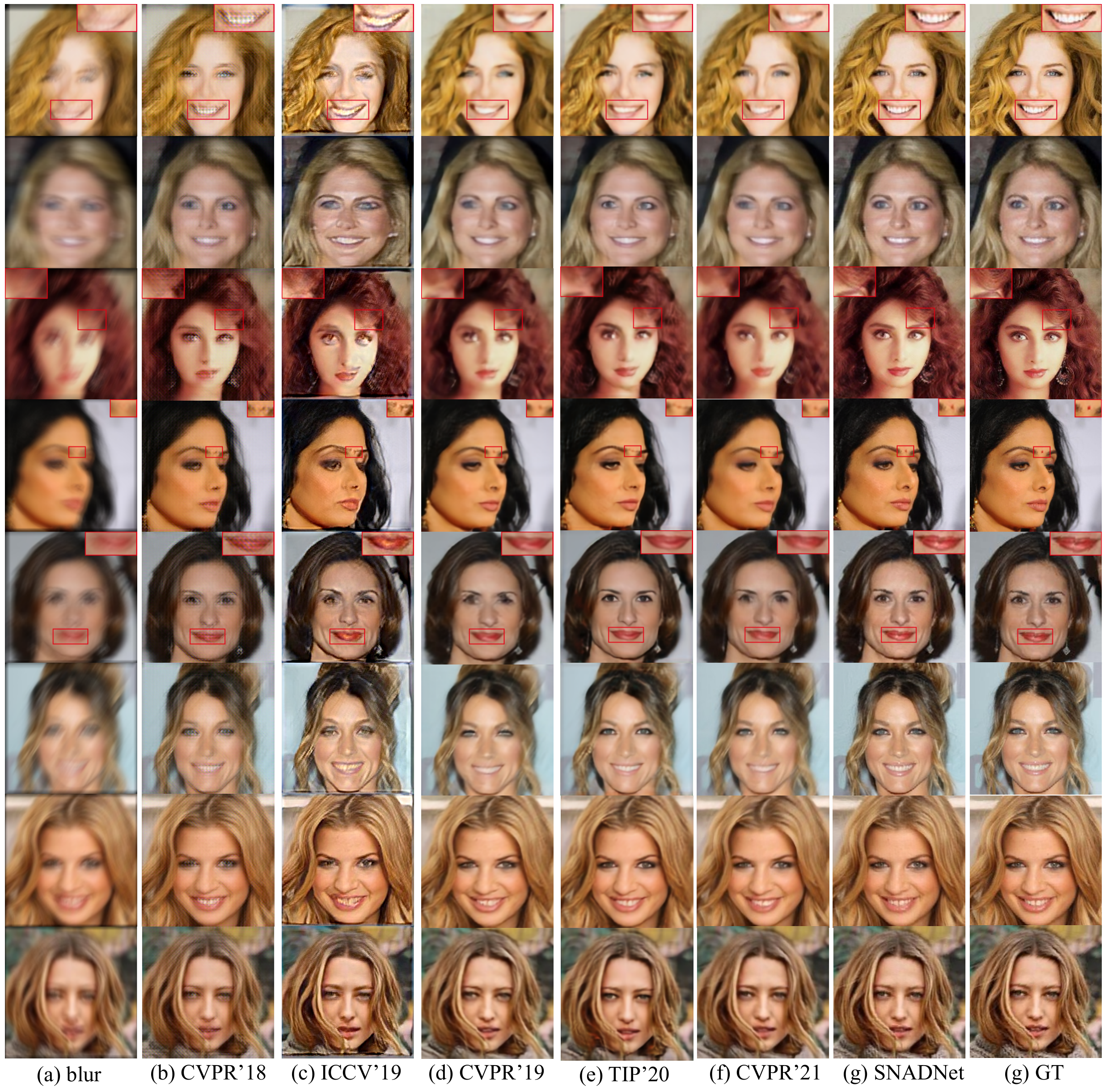}
     \caption{{Qualitative comparisons on the CelebA dataset. (a) blur face, (b) DeblurGAN,  (c) DeblurGAN-V2, (d) DMPHN, (e) UMSN, (f) MPRNet, and (g) SNADNet. More results are shown in the appendix.}}
    \label{fig_celeba}
\end{figure*}

\begin{table}
\centering
\caption{Comparison SNADNet with other advanced methods on CelebA-HQ. $\uparrow$ Higher is better. $\downarrow$ Lower is better.}
\begin{tabular}{l|l|l|l|l}
\hline
Methods  &  PSNR $\uparrow$ &SSIM $\uparrow$ & $L_{1}$(\%) $\downarrow$ &FID $\downarrow$\\
\hline
blur           & 21.78  & 0.625 & 5.70       & 95.88 \\
DeblurGAN (CVPR'18)     & 26.29  & 0.816  &3.99        & 15.22 \\
DeblurGAN-V2 (ICCV'19)  & 25.27  & 0.789  & 4.14       & 26.11\\
DMPHN (CVPR'19)         & 25.98  & 0.814  & 3.60       & 19.89\\
UMSN (TIP'20)          & 27.29  & 0.842  &3.47        & 21.67\\
MPRNet (CVPR'21) &27.38 &0.845 &3.12 &31.27\\
SNADNet (ours)    & \textbf{29.25}  & \textbf{0.883}  & \textbf{2.79}     & \textbf{11.03}\\
\hline
\end{tabular}
\label{tab_hq_compare_other}
\end{table}

\begin{table}
\centering
\caption{{Comparison our SNADNet with other advanced methods on CelebA. $\uparrow$ Higher is better. $\downarrow$ Lower is better.}}
\begin{tabular}{l|l|l|l|l}
\hline
Methods  &  PSNR $\uparrow$ &SSIM $\uparrow$ & $L_{1}$(\%) $\downarrow$ &FID $\downarrow$\\
\hline
blur &21.78   &0.769   & 5.06      &99.17  \\
DeblurGAN (CVPR'18)     &23.47   & 0.753  &4.83        & 40.23 \\
DeblurGAN-V2 (ICCV'19)  &21.10   &0.75   &5.51       &74.81 \\
DMPHN (CVPR'19)         &28.33   &0.909   &2.44       &22.29\\
UMSN (TIP'20)          &27.80  &0.899    &2.75   & 30.14\\
MPRNet (CVPR'21) &29.26  &0.918  &2.25  &13.38\\
SNADNet (ours)            &\textbf{30.08}  &\textbf{0.945}   &\textbf{2.21}  &\textbf{3.25} \\
\hline
\end{tabular}
\label{tab_compare_other}
\vskip -0.1in
\end{table}

{\bf Qualitative Comparisons:} Fig.~\ref{fig_celeba_hq} and Fig.~\ref{fig_celeba} compare our approach to five state-of-the-art methods on high-quality CelebA-HQ and low-quality CelebA datasets, respectively. Note that we selected their best performance results shown in Fig.~\ref{fig_celeba_hq} and Fig.~\ref{fig_celeba}. We can see that although DeblurGAN and DeblurGAN-V2 perform well on CelebA-HQ with one kind size blur kernel, they deteriorate rapidly on CelebA with a more complex face blur. This illustrates that DeblurGAN and DeblurGAN-v2 capture semantic information using feature matching, but they did not consider pixel-level reconstruction. Therefore, it would lead to loss of face structure information and do not perform well on face detail reconstruction. DMPHN divides the image into multiple patches and employs MSE to complete face blur removal, which yields
over-smooth and unsharp images. UMSN uses the semantic label of the face as prior information to complete the face deblurring, it produces good results but its texture reconstruction is not meticulous. MPRNet uses multi-patch and multi-stage to complete the deblurring task, while it requires a large cost of resources. Moreover, as can be seen from the results of Fig.~\ref{fig_celeba_hq} and Fig.~\ref{fig_celeba}, this method is not efficient.
Both figures illustrate that our method achieves good performance than all the other methods, and generates clear and natural faces on both CelebA-HQ and CelebA.
Especially, five SOTA methods fail to deblurring tasks when dealing with the too blurry image (see the third row in Fig.~\ref{fig_celeba}).

{\bf Quantitative Comparisons:}
Four types of criteria is used to measure the performance of different methods: 1) peak signal-to-noise ratio (PSNR); 2) Structural SIMilarity (SSIM); 3) $L_1$ loss; 4) Frechet Inception Distance (FID). PSNR based on the error between pixels measures the degree of image distortion. SSIM evaluates image quality in terms of brightness, contrast, and structure. $L_1$ evaluates images at the pixel level. FID comprehensively represents the distance between the Inception feature vectors of the generated image and the real image in the same domain. FID determines the diversity of the sample distribution, but visually, the lower the FID value, the sharper the resulting image. PSNR, SSIM, and $L_1$ loss judge the similarity between the deblurring image and the target image in terms of the structure. The higher the value of PSNR and SSIM, the better the deblurring result, and the lower the value of $L_1$ loss, the better the deblurring result.

Table~\ref{tab_hq_compare_other} and Table~\ref{tab_compare_other} show the qualitative results of different methods on both CelebA-HQ and CelebA datasets, respectively.
We can observe that DeblurGAN and DeblurGAN-V2 achieve good results on the CelebA-HQ with single-scale face deblurring kernel, but have little effect on the CelebA with more complex face deblurring kernel. DMPHN and UMSN are significantly better than deblurGAN in terms of PSNR, SSIM, and $L_1$ loss, but are still significantly worse than our SNADNet on CelebA dataset. Moreover, the performance of MPRNet on PSNR, SSIM, and $L_1$ loss also worse than our method. The FID indicators of these SOTA methods are worse than our method, which indicates that the generated faces of their methods are not clear enough and their reconstruction of texture details is not fine enough. Our SNADNet achieves substantial improvement than all the other methods, in terms of PSNR, SSIM, $L_1$ and FID.

\begin{equation}
\label{adv_d_loss_all}
\begin{aligned}
    loss_{d}^{adv} = \lambda_{4}^{adv}*loss_{d4}^{adv}+\lambda_{3}^{adv}*loss_{d3}^{adv}\\
    +\lambda_{2}^{adv}*loss_{d2}^{adv}+\lambda_{1}^{adv}*loss_{d1}^{adv}
\end{aligned}
\end{equation}
where $loss_{d}^{adv}$ and $loss_{g}^{adv}$ have the same value in super parameter.

\begin{figure*}[t]
    \setlength{\abovecaptionskip}{-0.02cm}
   \setlength{\belowcaptionskip}{-0.3cm}
    \centering
    \includegraphics[width=1.0\textwidth]{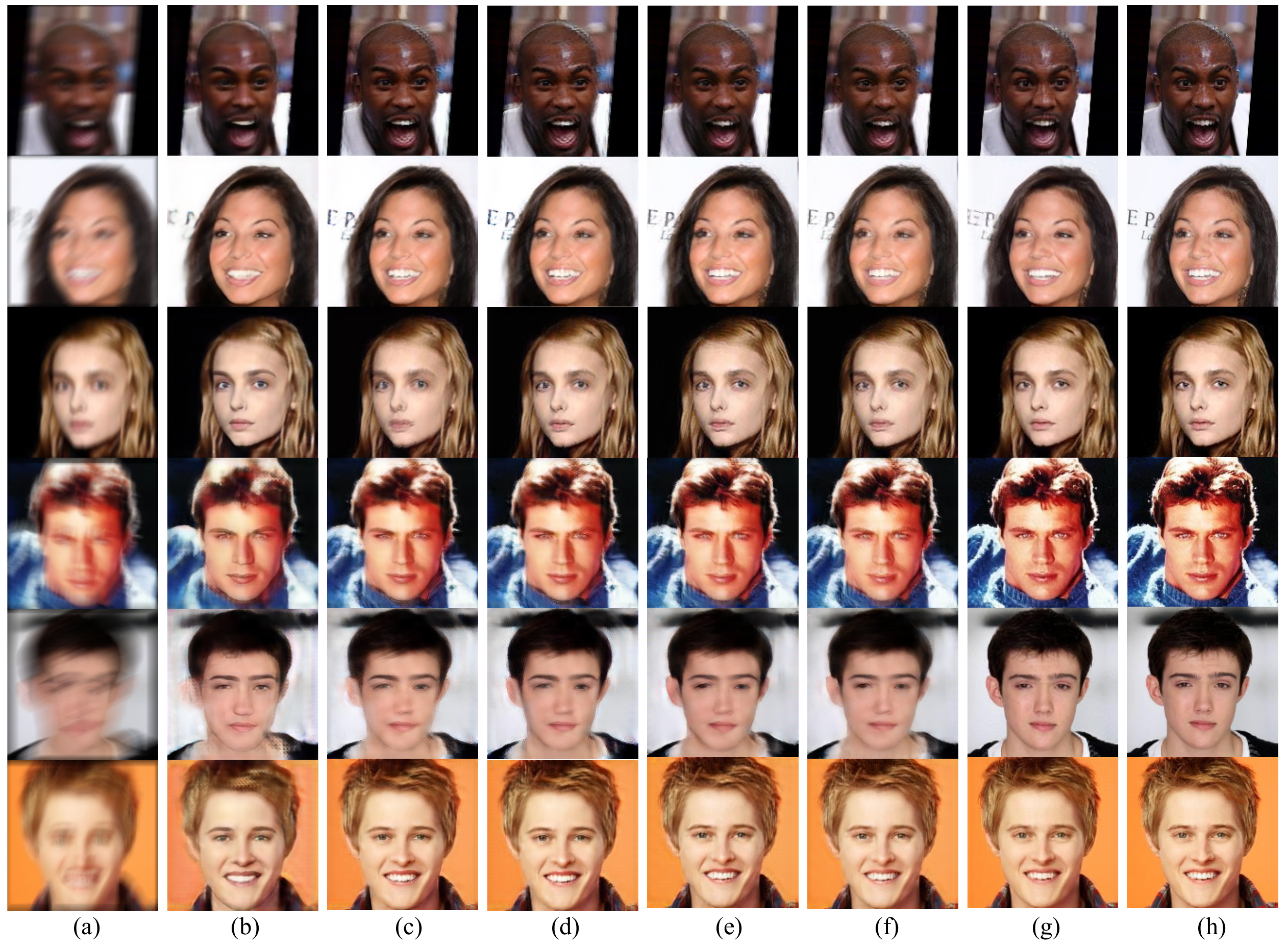}
     \caption{{Effect of different components in our model. (a) Blur face, (b) LinkNet, (c) LinkNet+BNAD, (d) LinKNet+INAD, (e) LinkNet+SNAD, (f) LinkNet+SNAD+TE+global discriminator,(g) LinkNet+SNAD+TE+Multi-discriminator (SNADNet), and (h) GT}}
    \label{fig_ablation}
\end{figure*}

\begin{table*}[h]
\centering
\caption{Quantitative deblurring results of our methods with different component on the CelebA-HQ dataset. Note that (b)-(g) match with (b)-(g) in Fig. \ref{fig_ablation}.}
\begin{tabular}{c|l|l|l|l}
\hline
Methods  &  PSNR $\uparrow$ &SSIM $\uparrow$ & $L_{1}$(\%) $\downarrow$ &FID $\downarrow$\\
\hline
(b) LinkNet+Multi-patch discriminator &25.34   & 0.865  &4.06  & 23.4\\
(c) LinkNet+BNAD+Multi-patch discriminator & 27.85  & 0.914  &  3.00      & 9.48\\
(d) LinKNet+INAD+Multi-patch discriminator & 28.26  & 0.921  & 2.84       &8.37  \\
(e) LinkNet+SNAD+Multi-patch discriminator   & 29.57  & 0.936  & 2.54       &  3.67\\
(f) LinkNet+SNAD+TE+global discriminator   & 29.68  & 0.938  & 2.52       &  3.45\\
(g) LinkNet+SNAD+TE+Multi-patch discriminator (SNADNet) & \textbf{30.08}  & \textbf{0.945}  & \textbf{2.21}       & \textbf{3.25}\\
\hline
\end{tabular}
\label{tab_ablation}
\vskip -0.25in
\end{table*}

\subsection{Ablation Study}
Our SNADNet is based on LinkNet, and we mainly designed SNAD and texture extractor (TE).
A Multi-patch discriminator is designed to supervise face generation. To further verify the function of each SNADNet module, we conducted ablation experiments to compare each module. Fig.~\ref{fig_ablation} illustrated that we mainly performed six comparative experiments: (a) blur image; (b) LinkNet+Multi patch discriminator, which only uses LinkNet as the generator and uses a multi-patch discriminator to supervise face generation; (c) LinNet+BNAD+Multi-patch discriminator, which consists of LinkNet, batch normalization and adaptive denormalization (BNAD), and multi-patch discriminator; (d) LinNet+INAD+Multi-patch discriminator, which replaces batch normalization (BN) with instance normalization (IN); (e) LinNet+SNAD+Multi-patch discriminator, which consists of LinkNet, separable normalization and adaptive denormalization (SNAD), and multi-patch discriminator; (f) LinNet+SNAD+TE+global discriminator, it not only includes SNAD, texture extractor (TE) employed but also uses global discriminator instead of multi-patch discriminator; (g) LinkNet+SNAD+TE+Multi-patch discriminator, it is our SNADNet; (h) Ground truth (GT).

Experimental results in terms of PSNR, SSIM, $L_1$ loss, and FID are shown in Table~\ref{tab_ablation}. Comparing (c), (d), and (e) with (b), we can see that that normalization and adaptive denormalization (AD) are effective. Comparing (e) with (c) and (d), we can know that separable normalization (SN) is better than batch normalization (BN) and instance normalization (IN) for face deblurring. Comparing (e) with (g), texture extractor (TE) makes the generated face much clear and more refined. Comparing (f) with (g), the multi-patch discriminator can better supervise the generator and generate higher quality faces, especially from the point of view of FID. From the ablation experiment, it can be known that SNADNet with linknet as the backbone, SNAD, and TE are adopted to constrain face structure and face texture details to generate high-quality faces, which is very effective for face deblurring.

\section{Conclusion}
In this paper, we proposed an effective SNADNet for Face deblurring.
SNADNet achieves substantial improvement by using separable normalization and adaptive denormalization (SNAD) and texture extractor (TE).
We designed separable normalization that can not only make the face foreground and background following the standard normal distribution, respectively but also make the whole feature following the standard normal distribution. Adaptive denormalization could regularize face structure and generate more harmonious and undistorted face structure, and a texture extractor (TE) could enhance facial texture and produce more detailed facial texture. Moreover, we designed a multi-patch discriminator to supervise our network to restore high-quality face images. Experiments show that our method has better performance than the state-of-the-art method.

\section*{Acknowledgment}
This work was supported by the Sichuan Science and Technology program (2019JDJQ0002, 2021016, 2020JDTD0020, 2019YFG0496) and National Science Foundation of China (42075142).

\bibliographystyle{IEEEtran}
\bibliography{bare_jrnl}

\begin{thebibliography}{10}
\providecommand{\url}[1]{#1}
\csname url@samestyle\endcsname
\providecommand{\newblock}{\relax}
\providecommand{\bibinfo}[2]{#2}
\providecommand{\BIBentrySTDinterwordspacing}{\spaceskip=0pt\relax}
\providecommand{\BIBentryALTinterwordstretchfactor}{4}
\providecommand{\BIBentryALTinterwordspacing}{\spaceskip=\fontdimen2\font plus
\BIBentryALTinterwordstretchfactor\fontdimen3\font minus
  \fontdimen4\font\relax}
\providecommand{\BIBforeignlanguage}[2]{{%
\expandafter\ifx\csname l@#1\endcsname\relax
\typeout{** WARNING: IEEEtran.bst: No hyphenation pattern has been}%
\typeout{** loaded for the language `#1'. Using the pattern for}%
\typeout{** the default language instead.}%
\else
\language=\csname l@#1\endcsname
\fi
#2}}
\providecommand{\BIBdecl}{\relax}
\BIBdecl

\bibitem{kupyn2018deblurgan}
O.~Kupyn, V.~Budzan, M.~Mykhailych, D.~Mishkin, and J.~Matas, ``Deblurgan:
  Blind motion deblurring using conditional adversarial networks,'' in
  \emph{Proceedings of the IEEE conference on computer vision and pattern
  recognition}, 2018, pp. 8183--8192.

\bibitem{kupyn2019deblurgan}
O.~Kupyn, T.~Martyniuk, J.~Wu, and Z.~Wang, ``Deblurgan-v2: Deblurring
  (orders-of-magnitude) faster and better,'' in \emph{Proceedings of the IEEE
  International Conference on Computer Vision}, 2019, pp. 8878--8887.

\bibitem{zhang2019deep}
H.~Zhang, Y.~Dai, H.~Li, and P.~Koniusz, ``Deep stacked hierarchical
  multi-patch network for image deblurring,'' in \emph{Proceedings of the IEEE
  Conference on Computer Vision and Pattern Recognition}, 2019, pp. 5978--5986.

\bibitem{yasarla2020deblurring}
R.~Yasarla, F.~Perazzi, and V.~M. Patel, ``Deblurring face images using
  uncertainty guided multi-stream semantic networks,'' \emph{IEEE Transactions
  on Image Processing}, vol.~29, pp. 6251--6263, 2020.

\bibitem{zamir2021multi}
S.~W. Zamir, A.~Arora, S.~Khan, M.~Hayat, F.~S. Khan, M.-H. Yang, and L.~Shao,
  ``Multi-stage progressive image restoration,'' in \emph{Proceedings of the
  IEEE/CVF Conference on Computer Vision and Pattern Recognition}, 2021, pp.
  14\,821--14\,831.

\bibitem{keinert2019robust}
F.~Keinert, D.~Lazzaro, and S.~Morigi, ``A robust group-sparse representation
  variational method with applications to face recognition,'' \emph{IEEE
  Transactions on Image Processing}, vol.~28, no.~6, pp. 2785--2798, 2019.

\bibitem{chen2021multi}
X.~Chen, X.~Luo, J.~Weng, W.~Luo, H.~Li, and Q.~Tian, ``Multi-view gait image
  generation for cross-view gait recognition,'' \emph{IEEE Transactions on
  Image Processing}, vol.~30, pp. 3041--3055, 2021.

\bibitem{wu2018look}
W.~Wu, C.~Qian, S.~Yang, Q.~Wang, Y.~Cai, and Q.~Zhou, ``Look at boundary: A
  boundary-aware face alignment algorithm,'' in \emph{Proceedings of the IEEE
  conference on computer vision and pattern recognition}, 2018, pp. 2129--2138.

\bibitem{chang2017faceposenet}
F.-J. Chang, A.~Tuan~Tran, T.~Hassner, I.~Masi, R.~Nevatia, and G.~Medioni,
  ``Faceposenet: Making a case for landmark-free face alignment,'' in
  \emph{Proceedings of the IEEE International Conference on Computer Vision
  Workshops}, 2017, pp. 1599--1608.

\bibitem{wang2019adaptive}
X.~Wang, L.~Bo, and L.~Fuxin, ``Adaptive wing loss for robust face alignment
  via heatmap regression,'' in \emph{Proceedings of the IEEE/CVF international
  conference on computer vision}, 2019, pp. 6971--6981.

\bibitem{pan2016l_0}
J.~Pan, Z.~Hu, Z.~Su, and M.-H. Yang, ``$ l\_0 $-regularized intensity and
  gradient prior for deblurring text images and beyond,'' \emph{IEEE
  transactions on pattern analysis and machine intelligence}, vol.~39, no.~2,
  pp. 342--355, 2016.

\bibitem{pan2016blind}
J.~Pan, D.~Sun, H.~Pfister, and M.-H. Yang, ``Blind image deblurring using dark
  channel prior,'' in \emph{Proceedings of the IEEE Conference on Computer
  Vision and Pattern Recognition}, 2016, pp. 1628--1636.

\bibitem{zhang2021exposure}
Y.~Zhang, C.~Wang, S.~J. Maybank, and D.~Tao, ``Exposure trajectory recovery
  from motion blur,'' \emph{IEEE Transactions on Pattern Analysis and Machine
  Intelligence}, 2021.

\bibitem{asim2020blind}
M.~Asim, F.~Shamshad, and A.~Ahmed, ``Blind image deconvolution using deep
  generative priors,'' \emph{IEEE Transactions on Computational Imaging},
  vol.~6, pp. 1493--1506, 2020.

\bibitem{shen2018deep}
Z.~Shen, W.-S. Lai, T.~Xu, J.~Kautz, and M.-H. Yang, ``Deep semantic face
  deblurring,'' in \emph{Proceedings of the IEEE Conference on Computer Vision
  and Pattern Recognition}, 2018, pp. 8260--8269.

\bibitem{xu2017learning}
X.~Xu, D.~Sun, J.~Pan, Y.~Zhang, H.~Pfister, and M.-H. Yang, ``Learning to
  super-resolve blurry face and text images,'' in \emph{Proceedings of the IEEE
  international conference on computer vision}, 2017, pp. 251--260.

\bibitem{tao2018scale}
X.~Tao, H.~Gao, X.~Shen, J.~Wang, and J.~Jia, ``Scale-recurrent network for
  deep image deblurring,'' in \emph{Proceedings of the IEEE Conference on
  Computer Vision and Pattern Recognition}, 2018, pp. 8174--8182.

\bibitem{li2020efficient}
Y.~Li, M.~Tofighi, J.~Geng, V.~Monga, and Y.~C. Eldar, ``Efficient and
  interpretable deep blind image deblurring via algorithm unrolling,''
  \emph{IEEE Transactions on Computational Imaging}, vol.~6, pp. 666--681,
  2020.

\bibitem{cai2020dark}
J.~Cai, W.~Zuo, and L.~Zhang, ``Dark and bright channel prior embedded network
  for dynamic scene deblurring,'' \emph{IEEE Transactions on Image Processing},
  vol.~29, pp. 6885--6897, 2020.

\bibitem{li2020dynamic}
L.~Li, J.~Pan, W.-S. Lai, C.~Gao, N.~Sang, and M.-H. Yang, ``Dynamic scene
  deblurring by depth guided model,'' \emph{IEEE Transactions on Image
  Processing}, vol.~29, pp. 5273--5288, 2020.

\bibitem{goodfellow2014generative}
I.~J. Goodfellow, J.~Pouget-Abadie, M.~Mirza, B.~Xu, D.~Warde-Farley, S.~Ozair,
  A.~C. Courville, and Y.~Bengio, ``Generative adversarial nets,'' in
  \emph{NIPS}, 2014.

\bibitem{arjovsky2017wasserstein}
M.~Arjovsky, S.~Chintala, and L.~Bottou, ``Wasserstein generative adversarial
  networks,'' in \emph{International conference on machine learning}.\hskip 1em
  plus 0.5em minus 0.4em\relax PMLR, 2017, pp. 214--223.

\bibitem{mao2017least}
X.~Mao, Q.~Li, H.~Xie, R.~Y. Lau, Z.~Wang, and S.~Paul~Smolley, ``Least squares
  generative adversarial networks,'' in \emph{Proceedings of the IEEE
  international conference on computer vision}, 2017, pp. 2794--2802.

\bibitem{johnson2016perceptual}
J.~Johnson, A.~Alahi, and L.~Fei-Fei, ``Perceptual losses for real-time style
  transfer and super-resolution,'' in \emph{European conference on computer
  vision}.\hskip 1em plus 0.5em minus 0.4em\relax Springer, 2016, pp. 694--711.

\bibitem{nah2017deep}
S.~Nah, T.~Hyun~Kim, and K.~Mu~Lee, ``Deep multi-scale convolutional neural
  network for dynamic scene deblurring,'' in \emph{Proceedings of the IEEE
  conference on computer vision and pattern recognition}, 2017, pp. 3883--3891.

\bibitem{song2019joint}
Y.~Song, J.~Zhang, L.~Gong, S.~He, L.~Bao, J.~Pan, Q.~Yang, and M.-H. Yang,
  ``Joint face hallucination and deblurring via structure generation and detail
  enhancement,'' \emph{International Journal of Computer Vision}, vol. 127, no.
  6-7, pp. 785--800, 2019.

\bibitem{ioffe2015batch}
S.~Ioffe and C.~Szegedy, ``Batch normalization: Accelerating deep network
  training by reducing internal covariate shift,'' in \emph{International
  conference on machine learning}.\hskip 1em plus 0.5em minus 0.4em\relax PMLR,
  2015, pp. 448--456.

\bibitem{wu2018group}
Y.~Wu and K.~He, ``Group normalization,'' in \emph{Proceedings of the European
  conference on computer vision (ECCV)}, 2018, pp. 3--19.

\bibitem{ulyanov2016instance}
D.~Ulyanov, A.~Vedaldi, and V.~Lempitsky, ``Instance normalization: The missing
  ingredient for fast stylization,'' \emph{arXiv preprint arXiv:1607.08022},
  2016.

\bibitem{ba2016layer}
J.~L. Ba, J.~R. Kiros, and G.~E. Hinton, ``Layer normalization,'' \emph{stat},
  vol. 1050, p.~21, 2016.

\bibitem{lin2017focal}
T.-Y. Lin, P.~Goyal, R.~Girshick, K.~He, and P.~Doll{\'a}r, ``Focal loss for
  dense object detection,'' in \emph{Proceedings of the IEEE international
  conference on computer vision}, 2017, pp. 2980--2988.

\bibitem{chaurasia2017linknet}
A.~Chaurasia and E.~Culurciello, ``Linknet: Exploiting encoder representations
  for efficient semantic segmentation,'' in \emph{2017 IEEE Visual
  Communications and Image Processing (VCIP)}.\hskip 1em plus 0.5em minus
  0.4em\relax IEEE, 2017, pp. 1--4.

\bibitem{liu2020rethinking}
H.~Liu, B.~Jiang, Y.~Song, W.~Huang, and C.~Yang, ``Rethinking image inpainting
  via a mutual encoder-decoder with feature equalizations,'' \emph{arXiv
  preprint arXiv:2007.06929}, 2020.

\bibitem{iizuka2017globally}
S.~Iizuka, E.~Simo-Serra, and H.~Ishikawa, ``Globally and locally consistent
  image completion,'' \emph{ACM Transactions on Graphics (ToG)}, vol.~36,
  no.~4, pp. 1--14, 2017.

\bibitem{zhou2020learning}
T.~Zhou, C.~Ding, S.~Lin, X.~Wang, and D.~Tao, ``Learning oracle attention for
  high-fidelity face completion,'' in \emph{Proceedings of the IEEE/CVF
  Conference on Computer Vision and Pattern Recognition}, 2020, pp. 7680--7689.

\bibitem{jolicoeur2018relativistic}
A.~Jolicoeur-Martineau, ``The relativistic discriminator: a key element missing
  from standard gan,'' \emph{arXiv preprint arXiv:1807.00734}, 2018.

\bibitem{jolicoeur2018rfdiv}
------, ``On relativistic f-divergences,'' \emph{arXiv preprint
  arXiv:1901.02474}, 2019.

\bibitem{karras2017progressive}
T.~Karras, T.~Aila, S.~Laine, and J.~Lehtinen, ``Progressive growing of gans
  for improved quality, stability, and variation,'' \emph{arXiv preprint
  arXiv:1710.10196}, 2017.

\bibitem{liu2015faceattributes}
Z.~Liu, P.~Luo, X.~Wang, and X.~Tang, ``Deep learning face attributes in the
  wild,'' in \emph{Proceedings of International Conference on Computer Vision
  (ICCV)}, December 2015.

\bibitem{boracchi2012modeling}
G.~Boracchi and A.~Foi, ``Modeling the performance of image restoration from
  motion blur,'' \emph{IEEE Transactions on Image Processing}, vol.~21, no.~8,
  pp. 3502--3517, 2012.

\end{thebibliography}

\appendices
\section{Normalization problem}
\label{nor-problem}

To illustrate the normalization problem, we denote $[B, C, H, W]$ as the shape of the feature map, {where $B$ is the batch size, $C$ is the number of the channel, $H$ is the height of the feature map, $W$ is the width of the image}. Since BN normalizes the feature map along the batch dimension, it computes mean and variance along the $[B, H, W]$ axes. Thus, BN is affected by the batch size. LN avoids batch dimension and normalize feature map along with the $[C, H, W]$ axes.  IN normalized dimension is $[H,W]$. GN is between LN and IN. First, the channel is divided into many groups, and each group is normalized. Furthermore, the dimensions of the feature are firstly reduced from $[N, C, H, W]$ reshape to $[N*G, C/G, H, W]$, and the normalized dimensions are $[C/G, H, W]$, {where $G$ is the number of groups and $C/G$ is the number of channels per group.} However, directly using these normalization methods for face data leads to a large amount of normalization bias.

An example is as follows to illustrate the problem. As shown in Fig.~\ref{fig_network} (a), we divide face $P_t$ into foreground region mask $M_f$ and background region mask $M_b$, and suppose $n_{f}$ and $n_{b}$ are their pixel numbers, respectively and set the number of face pixels $n=n_{f}+n_{b}$. Specifically, assume that $\mu$, $\mu_{f}$, $\mu_{b}$ and $\sigma^2$, $\sigma^2_f$, $\sigma^2_b$ are the mean and variance of the whole face feature, foreground and background, respectively.
Then, it holds that
\begin{equation}
\label{norm}
\begin{aligned}
   & \mu =\frac{n_{b}}{n}\mu_{b}+\frac{n_{f}}{n}\mu_{f}\\
   & \sigma ^{2}=\frac{n_{b}}{n}\sigma^{2}_{b}+\frac{n_{f}}{n}\sigma^{2}_{f}+\frac{n_{f}*n_{b}}{n^{2}}(\mu_{b}-\mu_{f})^{2}\\
\end{aligned}
\end{equation}
Since $n=n_{f}+n_{b}$, and  $\mu$, $\mu_{f}$, and $\mu_{b}$ are means of the whole face feature, foreground and background, respectively. It holds

\begin{equation}
\label{app1}
\begin{aligned}
     \mu =\frac{n_{b}}{n}\mu_{b}+\frac{n_{f}}{n}\mu_{f}
\end{aligned}
\end{equation}

{Since the variance of the face foreground is:}
\begin{equation}
\label{app2}
\begin{aligned}
     \sigma^{2}_{f}=E(x_f^2)-(E(x_f))^2=E(x_f^2)-\mu_{f}^2
\end{aligned}
\end{equation}

{It holds}
\begin{equation}
\label{app3}
\begin{aligned}
     E(x_f^2)=\sigma^{2}_{f}+\mu_{f}^2
\end{aligned}
\end{equation}
Similarly, for the background region, it holds
\begin{equation}
\label{app4}
\begin{aligned}
     E(x_b^2)=\sigma^{2}_{b}+\mu_{b}^2
\end{aligned}
\end{equation}

Based on the above Eq.(\ref{app1})-(\ref{app4}), the variance of the whole feature can be rewritten as follows.
\begin{equation}
\label{app5}
\begin{aligned}
      \sigma^{2}=&E(x^{2})-\mu^{2}\\
     &=\frac{\sum x^2}{n}-(\frac{n_b}{n}\mu_b+\frac{n_f}{n})^{2}\\
     & = \frac{n_b*(\sum x_{b}^2)}{n*n_b}+\frac{n_f*(\sum x_{f}^2)}{n*n_f}\\
     &-\frac{n_b^2}{n^2}\mu_b^2-\frac{n_f^2}{n^2}\mu_f^2-2*\frac{n_{b}n_{f}}{n^2}\mu_{f}\mu_{b}\\
    & =\frac{n_b}{n}E(x_b^2)+\frac{n_f}{n}E(x_f^2)\\
    & -\frac{n_b^2}{n^2}\mu_b^2-\frac{n_f^2}{n^2}\mu_f^2-2*\frac{n_{b}n_{f}}{n^2}\mu_{f}\mu_{b}\\
\end{aligned}
\end{equation}

Substitute Eq.(\ref{app3}) and Eq.(\ref{app4}) into Eq.(\ref{app5}), then
\begin{equation}
\label{app6}
\begin{aligned}
     \sigma^2&=\frac{n_f}{n}(\sigma_f^2+\mu_f^2)+\frac{n_b}{n}(\sigma_b^2+\mu_b^2)\\
     &-\frac{n_b^2}{n^2}\mu_b^2-\frac{n_f^2}{n^2}\mu_f^2-2*\frac{n_{b}n_{f}}{n^2}\mu_{f}\mu_{b}\\
    & =\frac{n_b(n-n_b)}{n^2}\mu_b^2+\frac{n_f(n-n_f)}{n^2}\mu_f^2\\
    & +\frac{n_b}{n}\sigma_b^2+\frac{n_f}{n}\sigma_f^2-2*\frac{n_{b}n_{f}}{n^2}\mu_{f}\mu_{b}\\
     &=\frac{n_f}{n}\sigma_f^2+\frac{n_b}{n}\sigma_b^2+
     \frac{n_{b}n_{f}}{n^2}(\mu_{b}^{2}+\mu_{f}^{2}-2*\mu_{b}\mu_{f})
\end{aligned}
\end{equation}

Therefore, 
\begin{equation}
\label{app7}
\begin{aligned}
     \sigma^2=
     \frac{n_f}{n}\sigma_f^2+\frac{n_b}{n}\sigma_b^2+
     \frac{n_{b}n_{f}}{n^2}(\mu_{b}-\mu_{f})^2
\end{aligned}
\end{equation}
We finished the proof.
From Eq.(\ref{norm}), we can know the relationships between $\mu$, $\mu_b$ and $\mu_f$, and that of  $\sigma^{2}$, $\sigma^{2}_f$ and $\sigma^{2}_b$. However, these is necessary and not sufficient conditions. If distributions of both foreground and background regions respectively follow the standard normal distribution, then the whole image also satisfies the standard normal distribution. But if the whole image follows the standard normal distribution, the foreground and background may not fit the standard normal distribution. Therefore, the conventional normalization method does not achieve the purpose of standard normal distribution of foreground and background respectively.
According to the definition of BN, GN, IN, and LN normalization, we can know that each position $x_{i}$ is recalculated. This can be seen from Eq.(\ref{norm2}) that both background and foreground will be normalized by means $\mu$ and variance $\sigma^{2}$ obtained from unified calculation. This would lead to a large amount of normalization bias for the face. In the following section, an approach is proposed to overcome this problem.

\begin{equation}
\label{norm2}
\begin{aligned}
   x_{i}^{'}=\frac{x_{i}-\mu }{\sqrt{\sigma^{2}+\xi }}
\end{aligned}
\end{equation}

\section{addition results.}
\begin{figure}[!h]
    \setlength{\abovecaptionskip}{-0.02cm}
   \setlength{\belowcaptionskip}{-0.3cm}
    \centering
    \begin{minipage}[t]{0.48\textwidth}
    \includegraphics[width=1.0\textwidth]{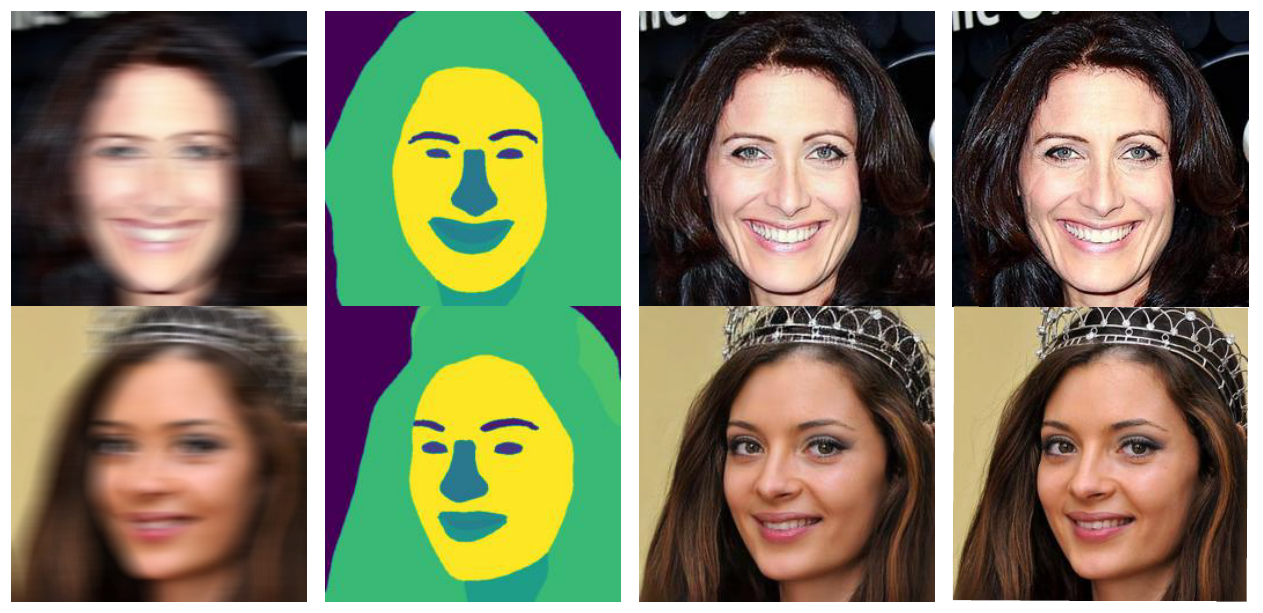}
    \end{minipage}
     \caption{{Results of face deblurring by SNADNet. The first column shows the blurry image. The second column consists of the corresponding semantic masks obtained from the face parsing network with fine-tuned blurry images. The third column shows the face deblurring results obtained using SNADNet. The fourth column is the target image.}}
    \label{appendix_more_result_parsing}
\end{figure}

\begin{figure}[!h]
    \setlength{\abovecaptionskip}{-0.02cm}
   \setlength{\belowcaptionskip}{-0.3cm}
    \centering
    \begin{minipage}[t]{0.48\textwidth}
    \includegraphics[width=1.0\textwidth]{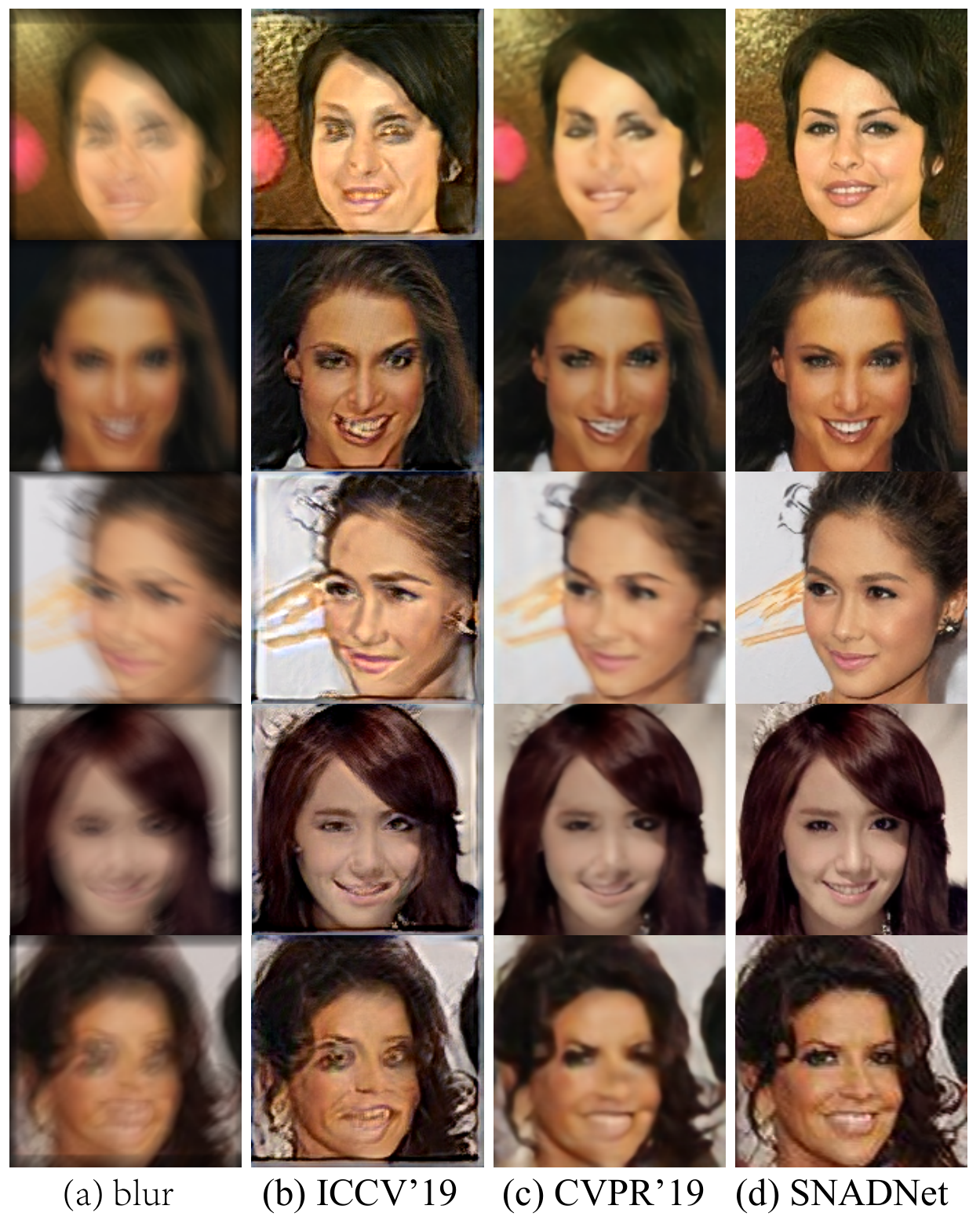}
    \end{minipage}
     \caption{{Qualitative comparisons with different SOTA methods. (a) blur, (b)DeblurGAN-V2, (c) DMPHN, (d) SNADNet (ours).}}
    \label{appendix_more_result_compare1}
\end{figure}

\begin{figure}[!h]
    \setlength{\abovecaptionskip}{-0.02cm}
   \setlength{\belowcaptionskip}{-0.3cm}
    \centering
    \begin{minipage}[t]{0.48\textwidth}
    \includegraphics[width=1.0\textwidth]{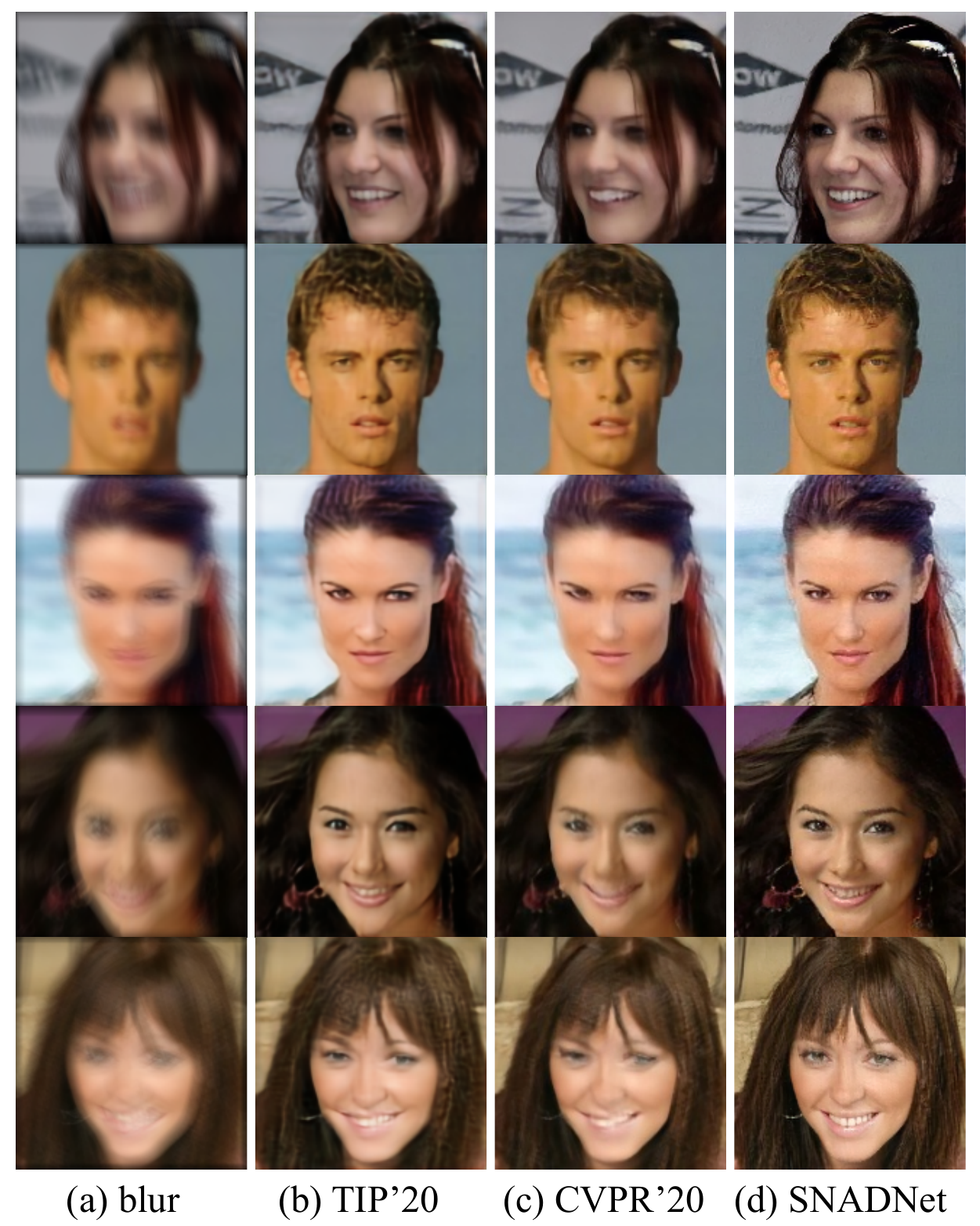}
    \end{minipage}
     \caption{{Qualitative comparisons of different SOTA methods. (a) blur, (b)UMSN, (c) MPRNet, (d) SNADNet (ours).}}
    \label{appendix_more_result_compare2}
\end{figure}

To further illustrate the effectiveness of our methods, more experimental results are added in this section. Specifically, Fig.~\ref{appendix_more_result_parsing} shows more corresponding semantic masks obtained from the face parsing network with fine-tuned on blurry face images and the deblurring face images by our SNADNet. Fig.~\ref{appendix_more_result_compare1} and Fig.~\ref{appendix_more_result_compare2} show more comparison results between the different SOTA networks and SNADNet.
Note that we split comparison results into two figures because of type setting.

\ifCLASSOPTIONcaptionsoff
  \newpage
\fi




\end{document}